\pdfoutput=1
\documentclass[11pt]{article}

\usepackage[]{acl}

\usepackage{times}
\usepackage{latexsym}
\usepackage{booktabs}
\usepackage{multirow}
\usepackage{graphicx}
\usepackage{subcaption}
\usepackage{amsfonts}  % for mathbb
\usepackage{enumitem}
\setlist[enumerate]{itemsep=0mm}
\newtheorem{definition}{Definition}
\newtheorem{problem}{Problem}

\newcommand{\llm}[0]{\textsc{llm}}
\newcommand{\llms}[0]{\textsc{llm}s}
\newcommand{\gnn}[0]{\textsc{gnn}}
\newcommand{\gnns}[0]{\textsc{gnn}s}
\newcommand{\grl}[0]{\textsc{grl}}

\usepackage[T1]{fontenc}
\usepackage[utf8]{inputenc}
\usepackage{microtype}
\usepackage{inconsolata}
\title{Understanding Survey Paper Taxonomy about Large Language Models via Graph Representation Learning}

\author{Jun Zhuang \\
  Department of Computer Science \\
  Boise State University \\
  \texttt{junzhuang@boisestate.edu} \\\And
  Casey Kennington \\
Department of Computer Science \\  
  Boise State University \\
  \texttt{caseykennington@boisestate.edu} \\}

\begin{document}
\maketitle

\begin{abstract}
As new research on Large Language Models (\llms) continues, it is difficult to keep up with new research and models. To help researchers synthesize the new research many have written survey papers, but even those have become numerous. In this paper, we develop a method to automatically assign survey papers to a taxonomy. We collect the metadata of 144 \llm\ survey papers and explore three paradigms to classify papers within the taxonomy. Our work indicates that leveraging graph structure information on co-category graphs can significantly outperform the language models in two paradigms; pre-trained language models' fine-tuning and zero-shot/few-shot classifications using \llms. We find that our model surpasses an average human recognition level and that fine-tuning \llms\ using weak labels generated by a smaller model, such as the GCN in this study, can be more effective than using ground-truth labels, revealing the potential of weak-to-strong generalization in the taxonomy classification task.
\end{abstract}
% TL;DR: We collected metadata about LLM surveys and developed a method for categorizing them into a taxonomy, indicating the superiority of graph representation learning over language models and revealing the efficacy of fine-tuning using weak labels.

\section{Introduction}
\label{sec:intro}
\begin{figure}[h]
  \centering
  \includegraphics[width=0.9\linewidth]{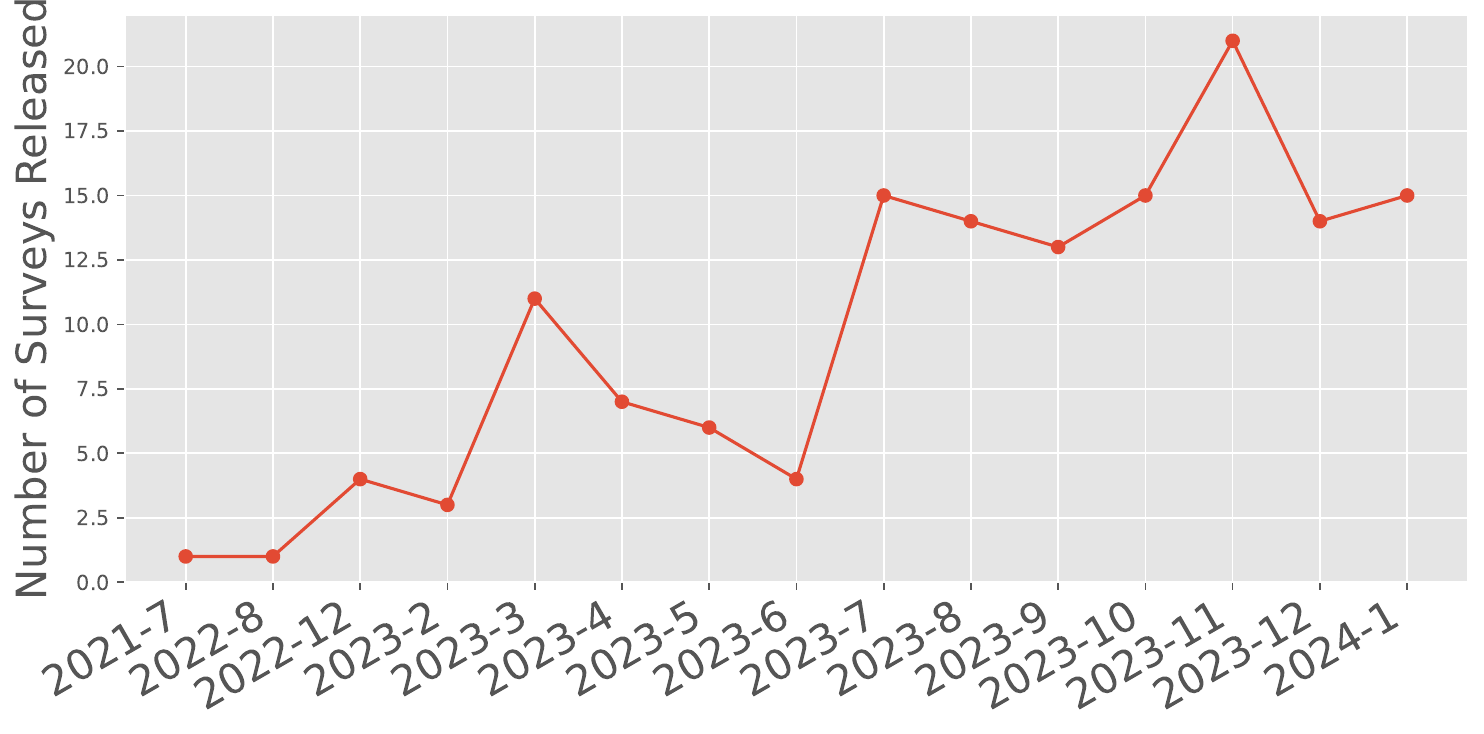}
  \caption{Trends of survey papers about large language models since 2021. Numbers reflect the year and month (e.g., 2023-3 is March 2023).}
\label{fig:trends}
\end{figure}

Collective attention in the field of Natural Language Processing (NLP)---and the wider public---has turned to Large Language Models (\llms). It has become so difficult to keep up with the proliferation of new models that many researchers have written survey papers to help synthesize the research progress. Survey papers are often crucial for newcomers to gain an in-depth understanding of the evolution of a research field. However, the volume of survey papers itself has become unruly for researchers---especially newcomers---to sift through. As illustrated in Figure~\ref{fig:trends}, the number of survey papers has been increasing significantly. This leads to our research question, aimed at aiding the field of NLP: Is it possible to automatically reduce the barriers for newcomers in a way that can keep up with the constant influx of new information?

In this paper, we address the above question by developing a method that can automatically assign survey papers to a taxonomy. Such a taxonomy will help researchers see new trends in the field and focus on specific survey papers that are relevant to their research. Classifying papers into a taxonomy may seem an ordinary task, but it is actually quite challenging for the following reasons:
\begin{enumerate}[itemsep=-2mm]
    \item Our dataset contains 144 papers. While this number for the survey papers is uncommonly large, the number of instances in the dataset is still relatively small.
    \item We propose a new taxonomy for the collected survey papers, where the distribution of each category is not uniform, which leads to a substantial class imbalance issue.
    \item Authors usually use similar terminologies to describe the \llms\ in the title and the abstract of these survey papers. Such textual similarity introduces significant difficulties in taxonomy classifications.
\end{enumerate}

To answer our research questions, we investigate three types of attributed graphs: text graphs, co-author graphs, and co-category graphs. Extensive experiments indicate that leveraging graph structure information of co-category graphs can help better classify the survey papers to the corresponding categories in the proposed taxonomy. Moreover, we validate that graph representation learning (\grl) can outperform language models in two paradigms, fine-tuning pre-trained language models and zero-shot/few-shot classifications using \llms.
Inspired by a recent study, which indicates that leveraging weak labels, which are generated by smaller (weaker) models, may help enhance the performance of larger (stronger) models~\cite{burns2023weak}, we further examine whether using weak labels, which are generated by \gnns\ in this study, in the fine-tuning paradigm can help the pre-trained language models. The experiments demonstrate that fine-tuning using weak labels can exceed that using ground-truth labels.
For the latter paradigm, we use the results of human recognition as the baseline. The analysis demonstrates that \grl\ achieves higher accuracy and F1 scores, and even surpasses the average human recognition level by a substantial margin.
Overall, our primary contributions can be summarized as follows:
\begin{itemize}[itemsep=-2mm]
  \item We collected and analyzed 144 survey papers about \llms\ and their metadata.\footnote{We will release all datasets and source code after this paper is accepted.}
  \item We propose a new taxonomy for categorizing the survey papers, which will be helpful for the research community, particularly newcomers and multidisciplinary research teams.
  \item Extensive experiments demonstrate that graph representation learning on co-category graph structure can effectively classify the papers and substantially outperform the language models and average human recognition level on a relatively small and class-imbalanced dataset with high textual similarity.
  \item Our results also reveal the potential of fine-tuning pre-trained language models using weak labels.
\end{itemize}

\section{Related Work}
\label{sec:rewk}

\paragraph{Taxonomy Classification}
Conventional taxonomy classification is a subset of Automatic Taxonomy Generation (ATG), which aims to generate taxonomy for a corpus~\cite{krishnapuram2003automatic}. The main challenge in ATG is to cluster the unlabeled topics into different hierarchical structures. Thus, most existing methods in ATG are clustering-based methods. \citet{zamir1998web} design a mechanism, Grouper, that dynamically groups and labels the search results. \citet{vaithyanathan1999model} propose a model to generate hierarchical clusters. \citet{lawrie2001finding} discover the relationship among words to generate concept hierarchies. Within these methods, a subset, called co-clustering, clusters keywords and documents simultaneously~\cite{frigui2002simultaneous, kummamuru2003fuzzy}.
Different from ATG, in this study, we classify survey papers into corresponding categories in the proposed taxonomy on relatively small and class-imbalanced datasets, whose text content contains similar terminologies.

\paragraph{Graph Representation Learning}
Graph representation learning (\grl) is a powerful approach for learning the representation in graph-structure data~\cite{zhou2020graph}, whereas most recent works achieve this goal using Graph Neural Networks (\gnns)~\cite{velivckovic2018graph, xu2018powerful}. \citet{bruna2013spectral} first introduce a significant advancement in convolution operations applied to graph data using both spatial method and spectral methods. To improve the efficiency of the eigendecomposition of the graph Laplacian matrix, \citet{defferrard2016convolutional} approximate spectral filters by using K-order Chebyshev polynomial. \citet{kipf2016semi} simplify graph convolutions to a first-order polynomial while yielding superior performance in semi-supervised learning. \citet{hamilton2017inductive} propose an inductive-learning approach that aggregates node features from corresponding fixed-size local neighbors. These \gnns\ have demonstrated exceptional performance in \grl, underscoring their significance in advancing this field.

\section{Methodology}
\label{sec:method}

In this section, we first introduce the procedure of data collection and then explore the metadata. We further explain the process of constructing three types of attributed graphs and how we learn graph representation via graph neural networks.

\subsection{Data Collection and Exploration}

\begin{figure}[h]
  \centering
  \includegraphics[width=0.9\linewidth]{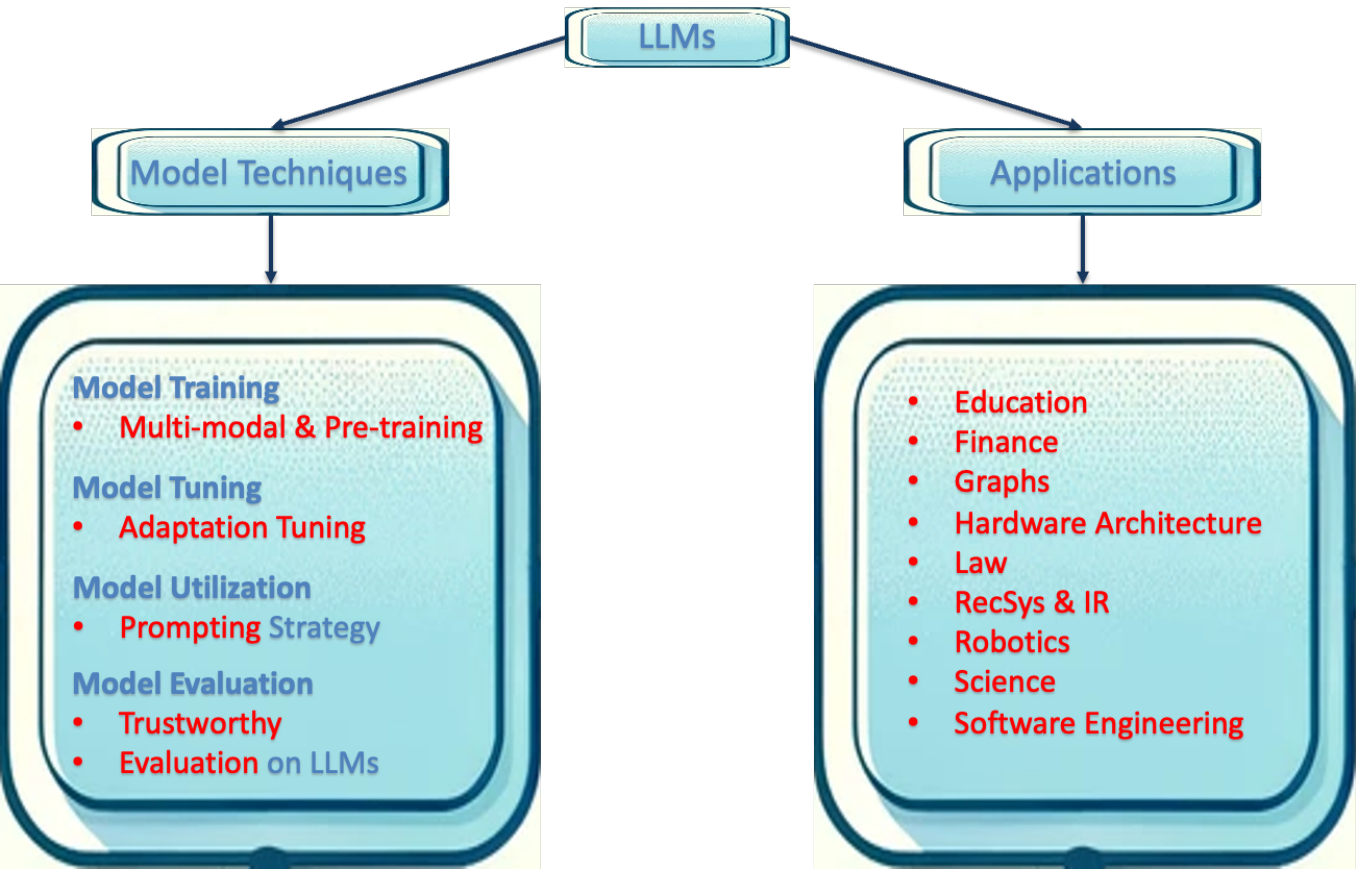}
  \caption{The mind map of survey papers about large language models. Besides "Comprehensive" and "Others" that are not included in the mind map, we highlight fourteen categories in our proposed taxonomy. The total number of categories for the 144 papers is sixteen.}
\label{fig:llm_mindmap}
\end{figure}

We scraped the metadata of survey papers about \llms\ from arXiv and further manually supplemented the metadata from Google Scholar and the ACL anthology. The papers range from July 2021 to January 2024. Given these survey papers, we designed a taxonomy and assigned each paper to a corresponding category within the taxonomy. Our motivation is that a reasonable taxonomy can provide a clear hierarchy of concepts for readers to better understand the relationship among a large number of survey papers. Though survey papers can be taxonomized differently, we noticed two broad categories: \emph{applications} and \emph{model techniques}. The \emph{applications} category further sub-divides into specific domains of focus (e.g., education or science), whereas \emph{model techniques} further sub-divides into ways of effecting models (e.g., fine-tuning).

\begin{figure}[h]
  \centering
  \includegraphics[width=0.9\linewidth]{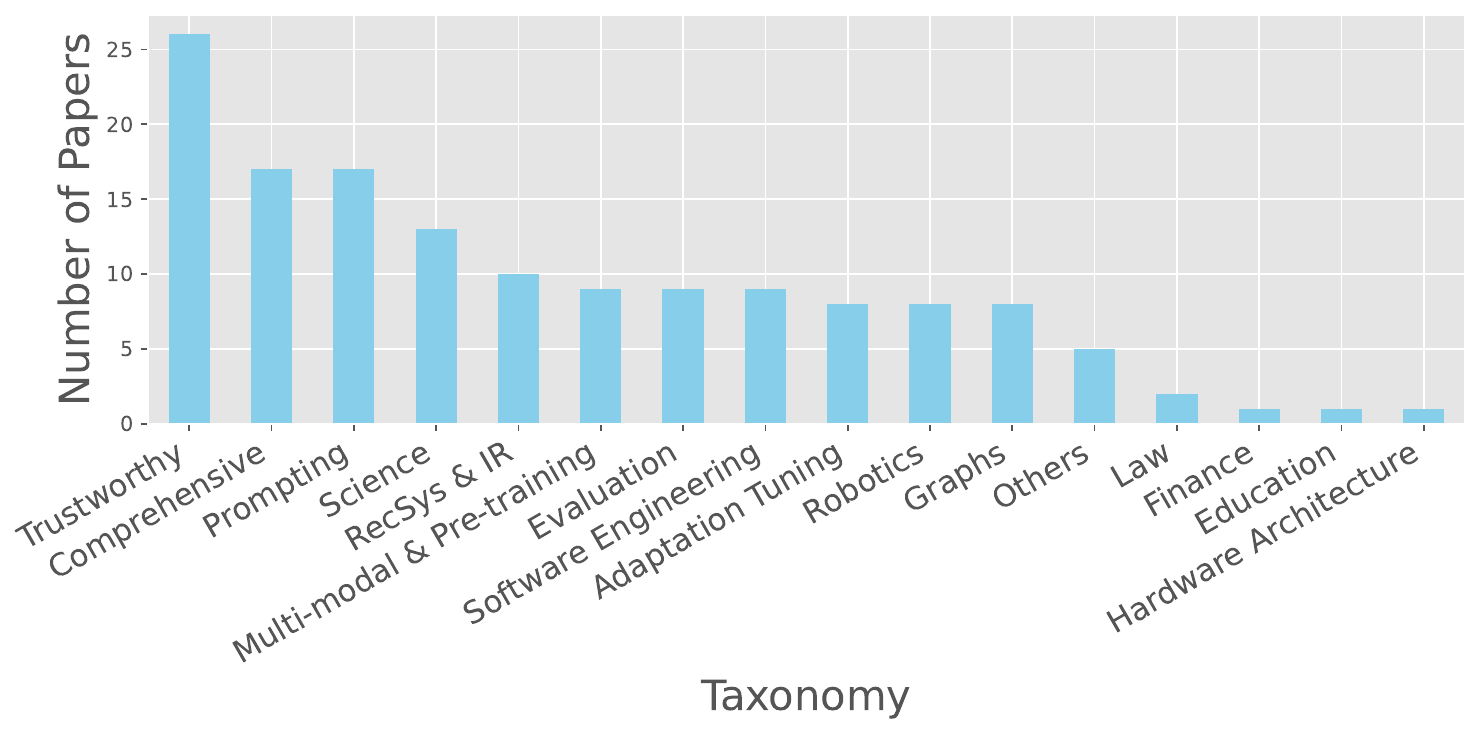}
  \caption{Distribution of classes in the taxonomy.}
\label{fig:tax_dist}
\end{figure}

We visualize our proposed taxonomy and highlight fourteen classes, i.e., the leaf nodes, in Figure~\ref{fig:llm_mindmap}. The total classes in the labels are sixteen, including \emph{comprehensive} and \emph{others} (not shown in the figure). To better understand the distribution of the classes, we present the class distribution in Figure~\ref{fig:tax_dist}. The distribution indicates that the class is extremely imbalanced, introducing a challenge to the taxonomy classification task.

\begin{figure}[h]
  \centering
  \includegraphics[width=0.8\linewidth]{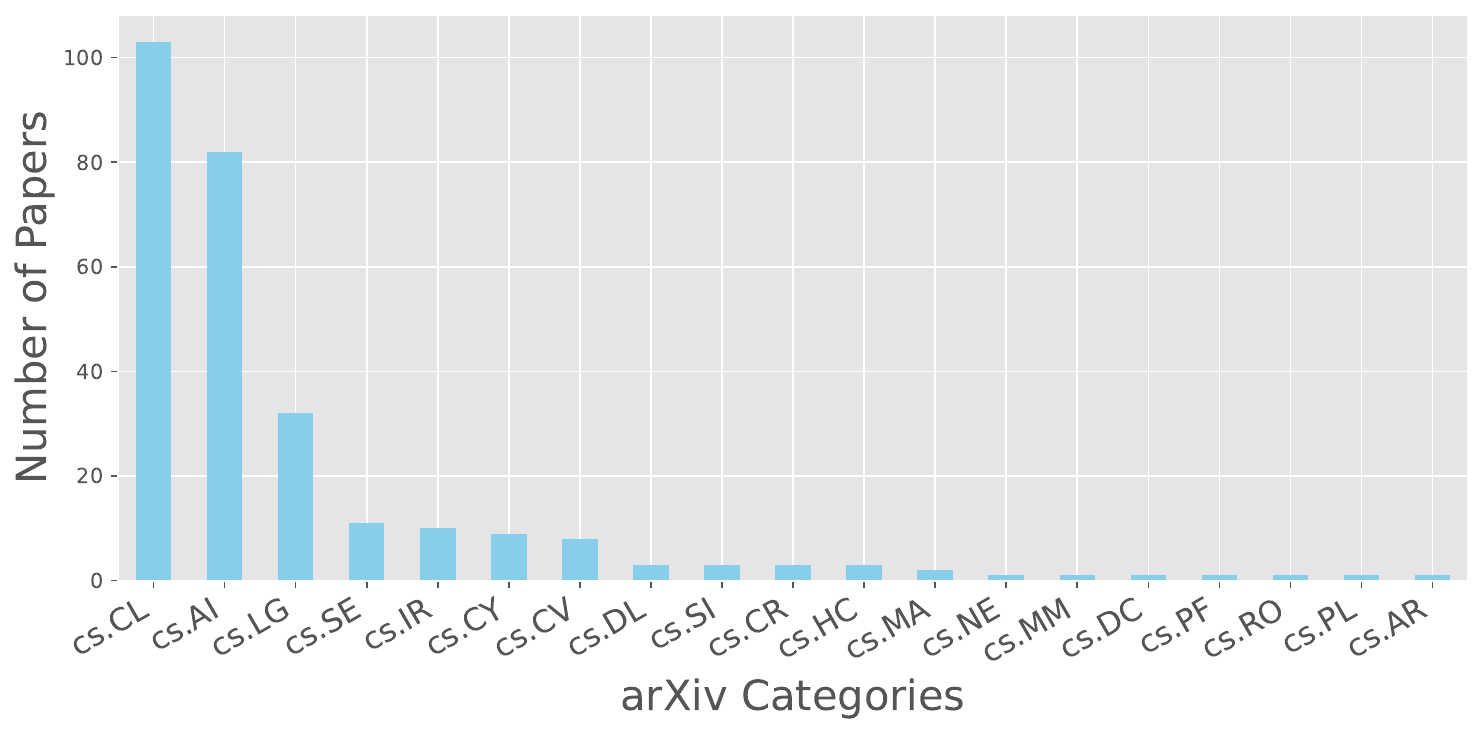}
  \caption{Distribution of survey papers that we found across different arXiv categories.}
\label{fig:cat_dist}
\end{figure}

After visualizing the proposed taxonomy, we further explain the motivation for proposing a new taxonomy instead of using the arXiv categories. In Figure~\ref{fig:cat_dist}, we present the distribution of survey papers across different arXiv categories. Top-2 frequent categories are \texttt{cs.CL} (Computation and Language), and \texttt{cs.AI} (Artificial Intelligence), which means that most authors choose these two categories for their works. However, these choices cannot help readers to better distinguish the survey papers. For example, papers related to model techniques are indistinguishable in arXiv categories. Thus, designing a new taxonomy is an essential step in this study.

\begin{figure*}[t]
  %\hfill
  \centering
  \begin{subfigure}{0.45\linewidth}
    \centering % include the first image
    \includegraphics[width=\textwidth]{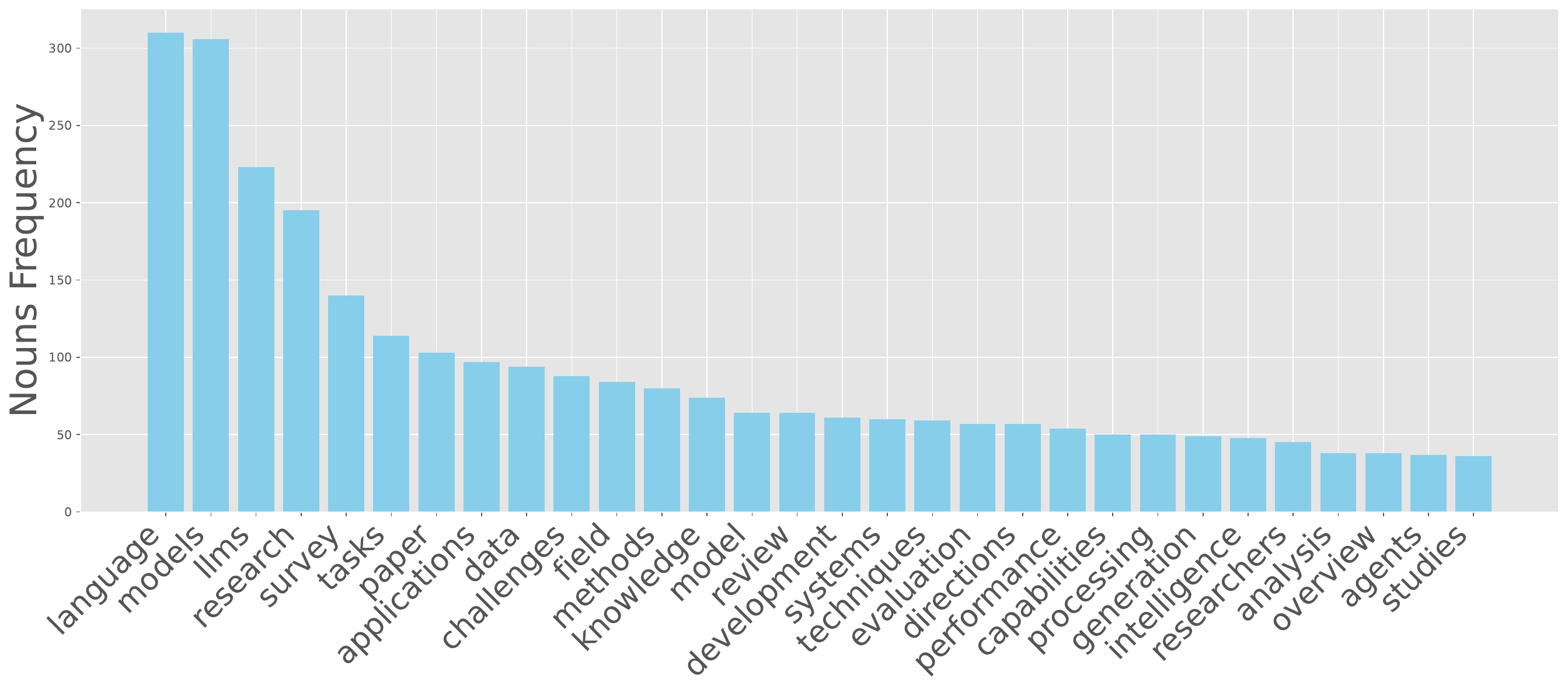}
  \end{subfigure}%
  %\hfill
  \begin{subfigure}{0.45\linewidth}
    \centering % include the second image
    \includegraphics[width=\textwidth]{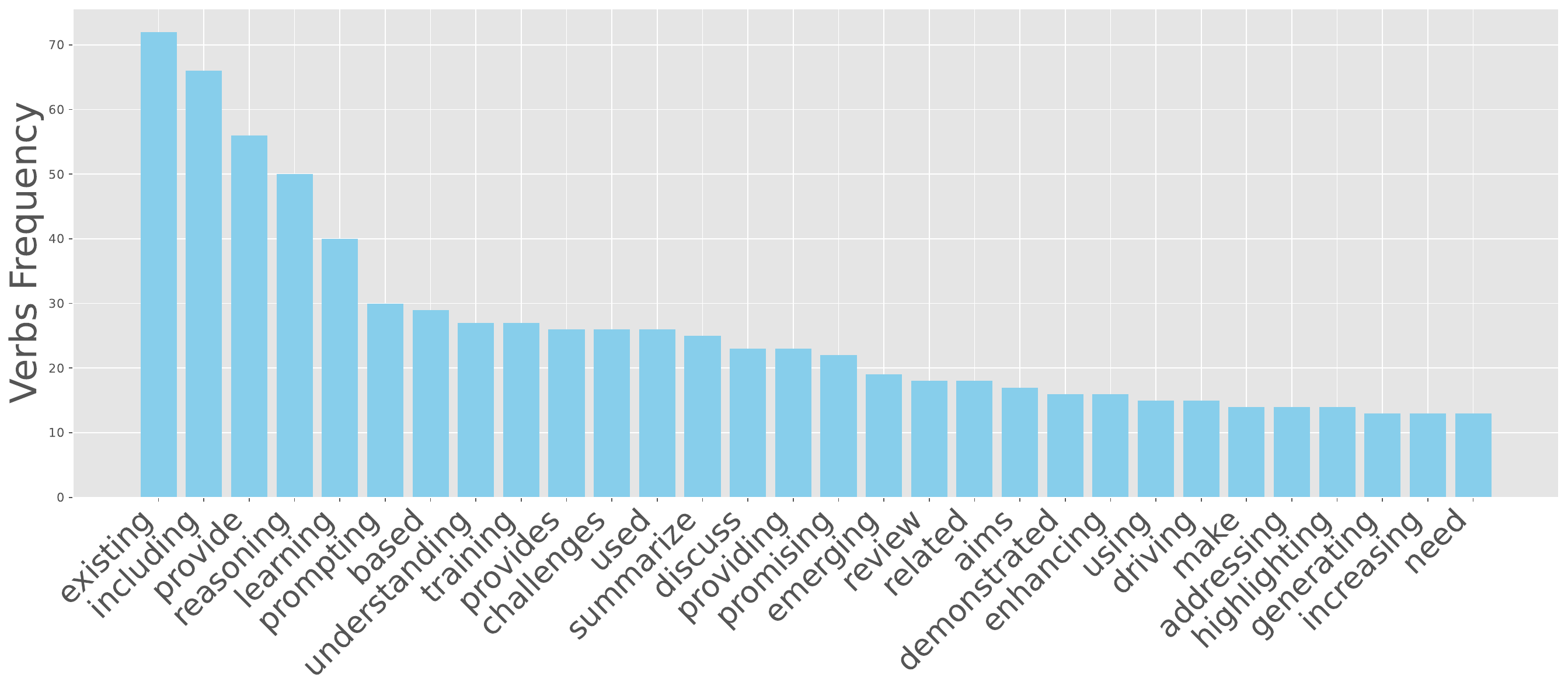}
  \end{subfigure}
\caption{Top 30 keywords frequency in the summary of survey papers.}
\label{fig:word_frq}
\end{figure*}

We also present the word frequency in Figure~\ref{fig:word_frq} to show which words have been frequently used in abstracts. These distributions suggest that the abstracts of these papers contain many similar terms, which increases the difficulty of text classification.

\begin{table}[h]
%\small
\scriptsize
\centering
\begin{tabular}{cc}
\toprule
{\bf Attributes} & {\bf Descriptions} \\
\midrule
{\bf Taxonomy} & Proposed taxonomy \\
{\bf Title} & Paper title \\
{\bf Authors} & Lists of author's name \\
{\bf Release Date} & First released date \\
{\bf Links} & Links of papers \\
{\bf Paper ID} & The arXiv paper ID \\
{\bf Categories} & The arXiv category \\
{\bf Summary} & Abstract of papers \\
\bottomrule
\end{tabular}
\caption{Data attributes and descriptions.}
\label{tab:data}
\end{table}

Overall, we present the data description of their attributes in Table~\ref{tab:data}. After designing the taxonomy and building the dataset, we explain how we classify documents into the taxonomy categories in the following section.

\subsection{Building Attributed Graphs}

The goal of building the graphs is to utilize the graph structure information to classify the taxonomy. Before building the graphs, We first define the attributed graphs as follows:
\begin{definition}
 An attributed graph $\mathcal{G}$ is a topological structure that represents the relationships among vertices associated with attributes. $\mathcal{G}$ consists of a set of vertices $\mathcal{V}$ = $\{ v_{1}, v_{2}, ..., v_{N} \}$ and edges $\mathcal{E} \subseteq \mathcal{V} \times \mathcal{V}$, where $N$ is the number of vertices in $\mathcal{G}$.
\label{dfn:attr_graph}
\end{definition}

Given the Definition~\ref{dfn:attr_graph}, we further define the matrix representation of an attributed graph as follows:
\begin{definition}
 Given an attributed graph $\mathcal{G}(\mathcal{V}, \mathcal{E})$, the topological relationship among vertices can be represented by a symmetric adjacency matrix $\mathbf{A} \in \mathbb{R}^{N \times N}$. Each vertex contains an attribute vector, a.k.a., a feature vector. All feature vectors constitute a feature matrix $\mathbf{X} \in \mathbb{R}^{N \times d}$, where $d$ is the number of features for each vertex. Thus, the matrix representation of an attributed graph can be formulated as $\mathcal{G}(\mathbf{A}, \mathbf{X})$.
\label{dfn:represent}
\end{definition}

Based on the above definitions, we build the graph by creating the term frequency-inverse document frequency (TF-IDF) feature matrices for both title and summary (i.e., abstract) columns, where the term frequency denotes the word frequency in the document, and inverse document frequency denotes the log-scaled inverse fraction of the number of documents containing the word. TF-IDF matrix is commonly used for text classification tasks because it helps capture the distinctive words that can indicate specific classes~\cite{yao2019graph}.
After establishing the TF-IDF matrices, we apply one-hot encoding on the arXiv’s categories and then combine three matrices along the feature dimension to build the feature matrix $\mathbf{X}$.
% 144 rows × 3065 columns

To leverage the topological information among vertices, we proceed to construct the graph structures to connect the attribute vectors. In this study, we are interested in three types of graphs: text graph, co-author graph, and co-category graph. We explain each type as follows.

\paragraph{Text Graph} We follow the same settings as TextGCN~\cite{yao2019graph} to build the text graph. Specifically, the edges of the text graph are built based on word occurrence (paper-word edges) in the paper's text data, including both title and summary, and word co-occurrence (word-word edges) in the whole text corpus. To obtain the global word co-occurrence information, we slide a fixed-size window on all papers' text data. Moreover, we calculate the edge weight between a paper vertex and a word vertex using the TF-IDF value of the word in the paper and calculate the edge weight between two-word vertices using point-wise mutual information (PMI), a popular metric to measure the associations between two words. Note that in the text graph, we don't use the above feature matrix because only paper vertices contain attribute vectors. To retain consistency, we set all values in the feature matrix as one. For the same reason, only the paper vertices are assigned labels, whereas all word vertices are labeled as a new class, which is not touched during the training or testing phase.

\paragraph{Co-author Graph} In the co-author graph, we introduce an edge connecting two vertices (papers) if they share at least one common author.

\paragraph{Co-category Graph} In the co-category graph, an edge is added between two vertices with at least one common arXiv category. In the co-authorship and co-category graphs, each vertex is assigned one class (taxonomy) as the label. Note that in this study all edges are undirected.

\subsection{Taxonomy Classification via Graph Representation Learning}

Given the well-built attributed graphs $\mathcal{G}(\mathbf{A}, \mathbf{X})$, we aim to investigate whether graph representation learning (\grl) using graph neural networks (\gnns) can help classify survey papers into the taxonomy. Before feeding the matrix representation, $\mathbf{A}$ and $\mathbf{X}$, of the attributed graphs $\mathcal{G}$ into \gnns, we first preprocess the adjacency matrix $\mathbf{A}$ as follows:
\begin{equation}
\mathbf{\hat{A}} = \mathbf{\tilde{D}}^{-\frac{1}{2}} \mathbf{\tilde{A}} \mathbf{\tilde{D}}^{-\frac{1}{2}},
\label{eqn:hat_a}
\end{equation}
where $\mathbf{\tilde{A}} = \mathbf{A} + I_{N}$, $\mathbf{\tilde{D}} = \mathbf{D} + I_{N}$. $I_{N}$ is an identity matrix. $\mathbf{D}_{i,i} = \sum_{j} \mathbf{A}_{i,j}$ is a diagonal degree matrix.

After preprocessing, we utilize \gnns\ to learn graph representation. The layer-wise message-passing mechanism of \gnns\ can be generally formulated as follows:
\begin{equation}
\textit{f}_{\mathbf{W}^{(l)}} \left(\mathbf{\hat{A}}, \mathbf{H}^{(l)} \right) = \sigma \left( \mathbf{\hat{A}} \mathbf{H}^{(l)} \mathbf{W}^{(l)} \right),
\label{eqn:mpm}
\end{equation}
where $\mathbf{H}^{(l)}$ is a node hidden representation in the $l$-th layer. The dimension of $\mathbf{H}^{(l)}$ in the input layer, middle layer, and output layer is the number of features $d$, hidden units $h$, and classes $K$, respectively. $\mathbf{H}^{(0)} = \mathbf{X}$. $\mathbf{W}^{(l)}$ is the weight matrix in the $l$-th layer. $\sigma$ denotes a non-linear activation function, such as ReLU.

In general node classification tasks, \gnns\ are trained with ground-truth labels $\mathbf{Y} \in \mathbb{R}^{N \times 1}$. In this study, we build the ground-truth labels based on our proposed taxonomy. To simplify the problem, each paper is assigned one primary category as the label, even if the paper sometimes may belong to more than one category. During training, we optimize \gnns\ with cross-entropy.

In brief, we address the Taxonomy Classification problem via \grl\ approaches in this study and formally state the problem as follows:
\begin{problem}
After building an attributed graph $\mathcal{G}(\mathbf{\hat{A}}, \mathbf{X})$ and the ground-truth labels $\mathbf{Y}$ based on the survey metadata, we train a graph neural network (\gnn) on the train data and evaluate the taxonomy classification performance on the test data. Our goal is to design a method to better understand (classify) the taxonomy of the survey papers.
\end{problem}

\section{Experiment}
\label{sec:exp}
In this section, we evaluate the graph representation learning (\grl)'s effectiveness compared with two paradigms using language models.

\paragraph{Experimental Settings}

\begin{table}[h] % Dataset
\footnotesize
%\scriptsize
\centering
\setlength{\tabcolsep}{3.8pt}
\begin{tabular}{cccccc}
  \toprule
    Subsets & Graphs & {$\left| \mathcal{V} \right|$} & {$\left| \mathcal{E} \right|$} & {$\left| F \right|$} & {$\left| C \right|$} \\
    \midrule
    \multirow{3}{*}{$Data_{Nov23}$} & Text & 737 & 94,943 & 737 & 16 \\
    & Co-author & 112 & 204 & 3,065 & 15 \\
    & Co-category & 112 & 4,908 & 3,065 & 15 \\
    \midrule
    \multirow{3}{*}{$Data_{Jan24}$} & Text & 951 & 137,709 & 951 & 17 \\
    & Co-author & 144 & 332 & 3,542 & 16 \\
    & Co-category & 144 & 8,140 & 3,542 & 16 \\
    \midrule
    \multirow{3}{*}{$Data_{subset}$} & Text & 905 & 128,575 & 905 & 12 \\
    & Co-author & 134 & 302 & 3,394 & 11 \\
    & Co-category & 134 & 6,964 & 3,394 & 11 \\ 
  \bottomrule
\end{tabular}
\caption{Statistics of graph datasets. $\left| \mathcal{V} \right|$, $\left| \mathcal{E} \right|$, $\left| F \right|$, and $\left| C \right|$ denote the number of nodes, edges, features, and classes, respectively.}
\label{table:dataset}
\end{table}

\begin{table*}[t]
\small
\centering
\setlength{\tabcolsep}{2.5pt}
\scalebox{0.9}{
\begin{tabular}{ccccccc}
\toprule
%\multicolumn{1}{c}{\multirow{2}{*}{\bf Models}} 
& \multicolumn{2}{c}{\bf $Data_{Nov23}$} & \multicolumn{2}{c}{\bf $Data_{Jan24}$} & \multicolumn{2}{c}{\bf $Data_{subset}$} \\
\cmidrule(lr){2-3} \cmidrule(lr){4-5} \cmidrule(lr){6-7}
& Accuracy & Weighted-F1 & Accuracy & Weighted-F1 & Accuracy & Weighted-F1 \\
\midrule
{\bf Text} & 20.91 (5.45) & 14.20 (4.41) & 17.86 (7.14) & 16.31 (4.49) & 23.08 (4.87) & 18.82 (1.50) \\
{\bf Co-author} & 33.04 (8.06) & 33.06 (8.69) & 20.00 (8.56) & 19.24 (8.79) & 29.63 (7.03) & 29.24 (6.02) \\
{\bf Co-category (All)} & 63.48 (18.36) & 62.82 (16.96) & 75.17 (5.52) & 74.60 (4.81) & {\bf 79.26 (6.87)} & {\bf 77.88 (7.52)} \\
{\bf Co-category (Rm \texttt{cs.CL})} & 70.43 (9.28) & 68.46 (9.63) & 67.59 (15.36) & 65.81 (17.03) & 76.30 (12.31) & 73.83 (14.53) \\
{\bf Co-category (Rm \texttt{cs.AI})} & {\bf 73.91 (18.03)} & {\bf 72.41 (18.28)} & {\bf 75.86 (8.99)} & {\bf 75.79 (9.62)} & 77.04 (3.63) & 74.15 (4.11) \\
{\bf Co-category (Rm \texttt{cs.CL}, \texttt{cs.AI})} & 26.09 (10.65) & 20.19 (10.56) & 37.93 (8.45) & 35.97 (7.92) & 49.63 (7.63) & 47.32 (7.59) \\
\midrule
{\bf Co-category (Rm \texttt{cs.IR})} & 63.48 (18.36) & 62.82 (16.96) & 75.17 (5.52) & 74.60 (4.81) & 79.26 (6.87) & 77.88 (7.52) \\
{\bf Co-category (Rm \texttt{cs.RO})} & 63.48 (18.36) & 62.82 (16.96) & 75.17 (5.52) & 74.60 (4.81) & 79.26 (6.87) & 77.88 (7.52) \\
{\bf Co-category (Rm \texttt{cs.SE})} & \textcolor{gray}{65.22 (11.00)} & \textcolor{gray}{63.21 (10.37)} & \textcolor{gray}{74.48 (8.33)} & \textcolor{gray}{75.10 (6.77)} & \textcolor{gray}{82.96 (6.02)} & \textcolor{gray}{82.93 (6.22)} \\
{\bf Co-category (Rm \texttt{cs.IR}, \texttt{cs.RO})} & 63.48 (18.36) & 62.82 (16.96) & 75.17 (5.52) & 74.60 (4.81) & 79.26 (6.87) & 77.88 (7.52) \\
{\bf Co-category (Rm \texttt{cs.IR}, \texttt{cs.SE})} & \textcolor{gray}{65.22 (11.00)} & \textcolor{gray}{63.21 (10.37)} & \textcolor{gray}{74.48 (8.33)} & \textcolor{gray}{75.10 (6.77)} & \textcolor{gray}{82.96 (6.02)} & \textcolor{gray}{82.93 (6.22)} \\
{\bf Co-category (Rm \texttt{cs.RO}, \texttt{cs.SE})} & \textcolor{gray}{65.22 (11.00)} & \textcolor{gray}{63.21 (10.37)} & \textcolor{gray}{74.48 (8.33)} & \textcolor{gray}{75.10 (6.77)} & \textcolor{gray}{82.96 (6.02)} & \textcolor{gray}{82.93 (6.22)} \\
{\bf Co-category (Rm \texttt{cs.IR}, \texttt{cs.RO}, \texttt{cs.SE})} & \textcolor{gray}{65.22 (11.00)} & \textcolor{gray}{63.21 (10.37)} & \textcolor{gray}{74.48 (8.33)} & \textcolor{gray}{75.10 (6.77)} & \textcolor{gray}{82.96 (6.02)} & \textcolor{gray}{82.93 (6.22)} \\
\bottomrule
\end{tabular}
}
\caption{Evaluation of graph representation learning on three types of attributed graphs across three subsets of our data. We also conducted ablation studies on the graph structure of co-category graphs by removing (denoted by Rm) some arXiv categories. We ran the experiments five times and presented the mean (std).}
\label{tab:gnn}
\end{table*}

\begin{figure*}[t]
  %\hfill
  \begin{subfigure}{0.245\linewidth}
    \centering  % include the 1st image
    \includegraphics[width=\textwidth]{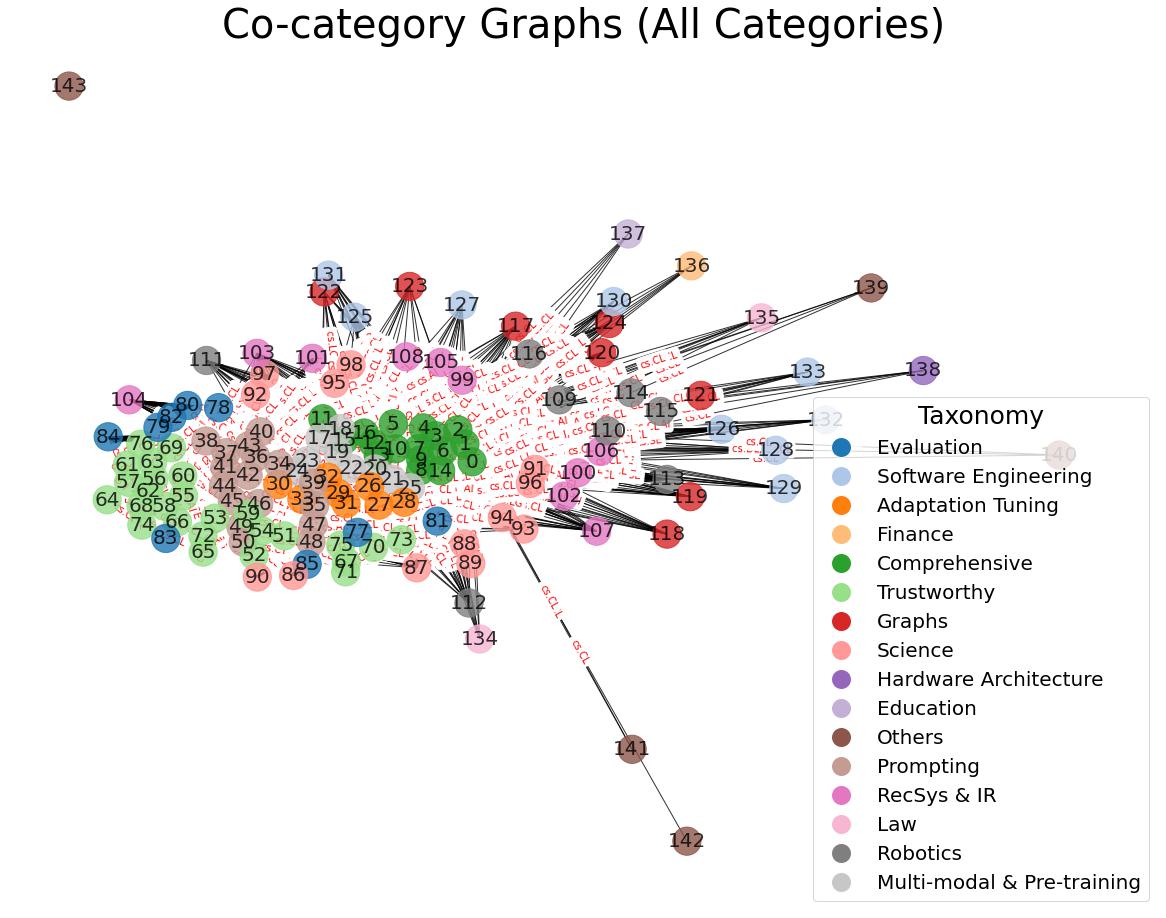}
  \end{subfigure}
  %\hfill
  \begin{subfigure}{0.245\linewidth}
    \centering  % include the 3rd image
    \includegraphics[width=\textwidth]{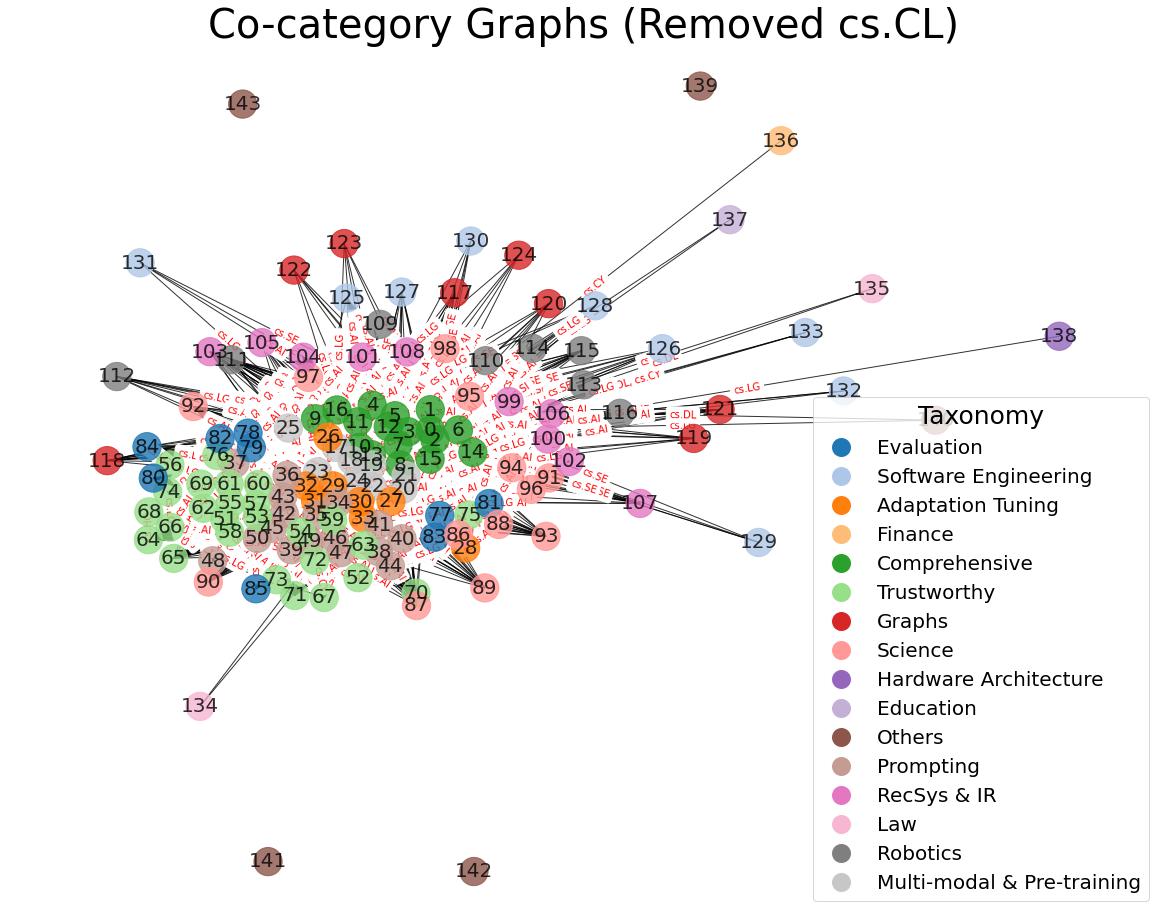}
  \end{subfigure}
  %\hfill
  \begin{subfigure}{0.245\linewidth}
    \centering  % include the 4th image
    \includegraphics[width=\textwidth]{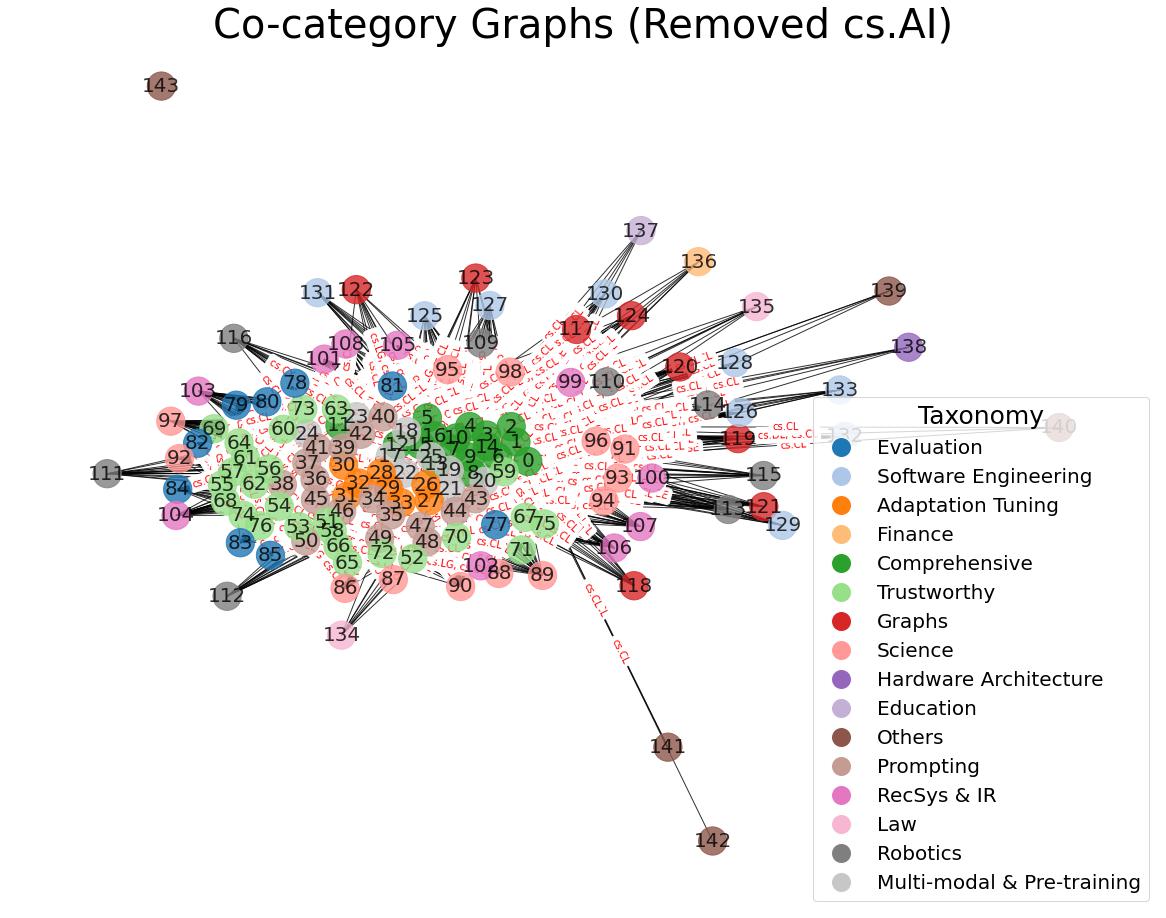}
  \end{subfigure}
  %\hfill
  \begin{subfigure}{0.245\linewidth}
    \centering  % include the 2nd image
    \includegraphics[width=\textwidth]{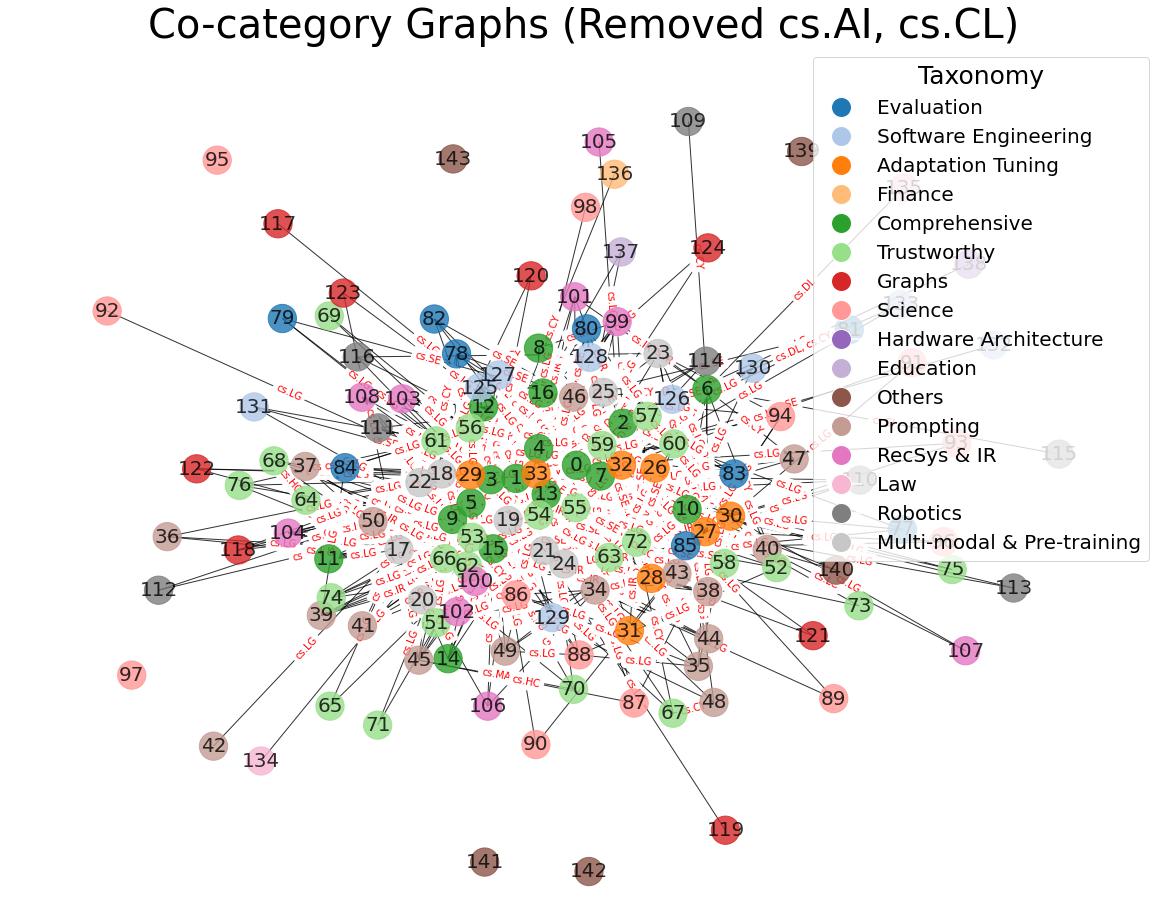}
  \end{subfigure}
\caption{Visualization of four graph structures on co-category graphs by removing the categories.}
\label{fig:cocat_graphs}
\end{figure*}

\begin{figure*}[t]
  %\hfill
  \begin{subfigure}{0.245\linewidth}
    \centering  % include the 1st image
    \includegraphics[width=\textwidth]{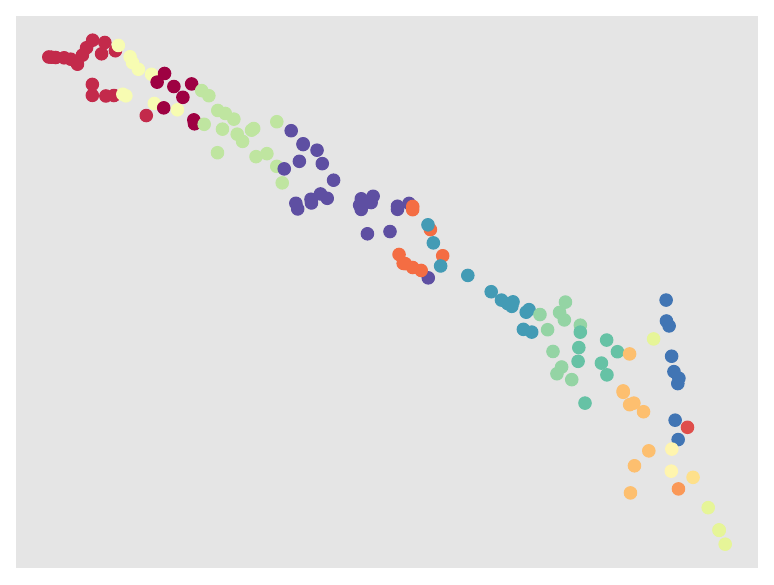}
    \caption{All Categories}
  \end{subfigure}
  %\hfill
  \begin{subfigure}{0.245\linewidth}
    \centering  % include the 3rd image
    \includegraphics[width=\textwidth]{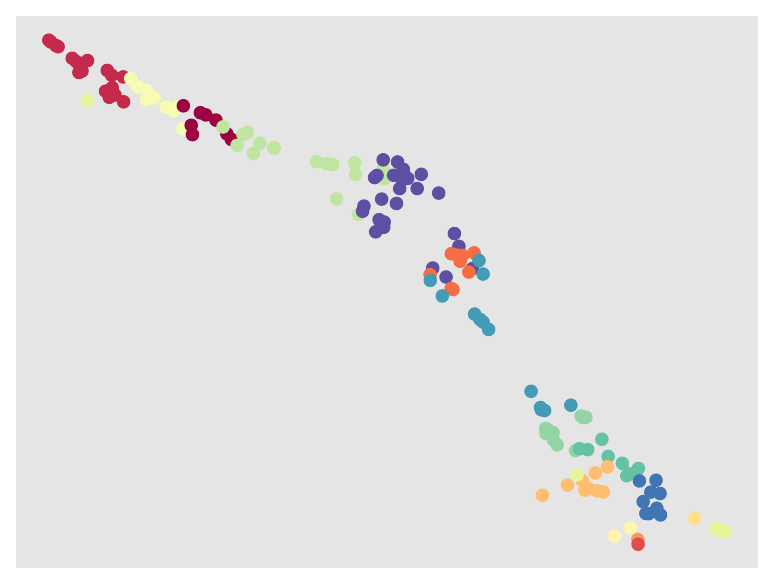}
    \caption{Removed \texttt{cs.CL}}
  \end{subfigure}
  %\hfill
  \begin{subfigure}{0.245\linewidth}
    \centering  % include the 4th image
    \includegraphics[width=\textwidth]{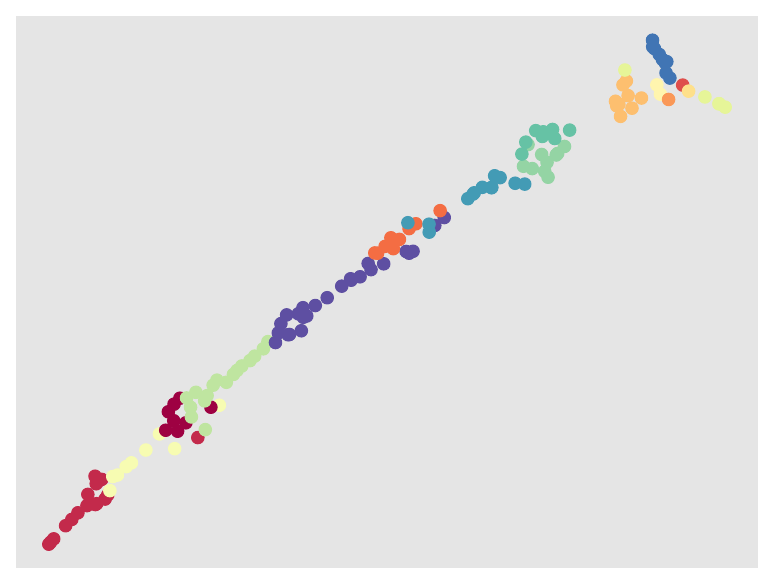}
    \caption{Removed \texttt{cs.AI}}
  \end{subfigure}
  %\hfill
  \begin{subfigure}{0.245\linewidth}
    \centering  % include the 2nd image
    \includegraphics[width=\textwidth]{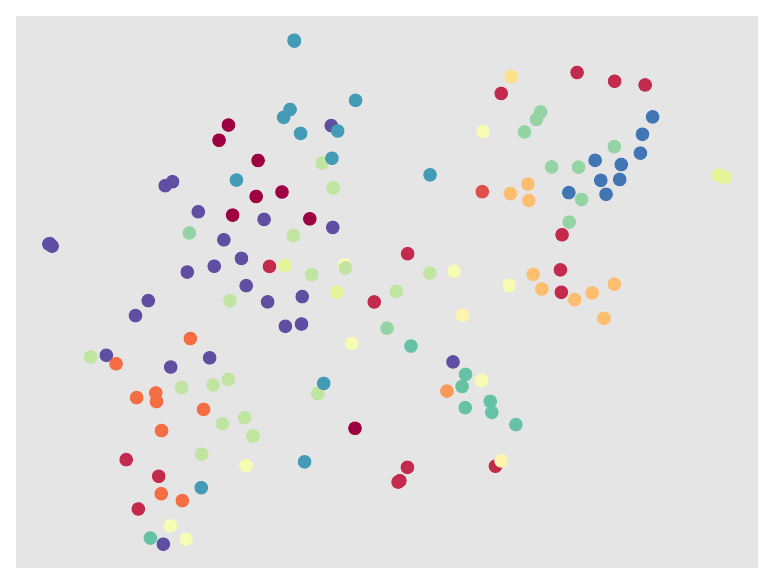}
    \caption{Removed \texttt{cs.CL}, \texttt{cs.AI}}
  \end{subfigure}
\caption{Visualization of GCNs' hidden representation on co-category graphs in $Data_{Jan24}$ via t-SNE. Each dot represents one node and is labeled with one color.}
\label{fig:cocat_embs}
\end{figure*}

To examine the generalization of our method on various graph structures, we investigate three types of attributed graphs: text graphs, co-author graphs, and co-category graphs, and compare the classification performance of \grl\ with that of fine-tuning pre-trained language models across three subsets of our data. Both $Data_{Nov 23}$ and $Data_{Jan 24}$ contain survey papers collected at the end of corresponding months (November 2023 and January 2024). $Data_{Jan 24}$ includes a new category, \emph{Hardware Architecture}. We further construct the third subset $Data_{subset}$ by removing some proposed categories with fewer instances in $Data_{Jan 24}$; these categories are \emph{Law}, \emph{Finance}, \emph{Education}, \emph{Hardware Architecture}, and \emph{Others}.
The motivation for constructing three subsets is to validate the generalization of our method across different subsets since the classification performance may significantly change on small datasets. Also, new categories may emerge at any period because research on \llms\ is developing rapidly, and so are related survey papers. For example, a new category, \emph{Hardware Architecture}, emerges in $Data_{Jan 24}$. The change of categories may affect the performance as well. Therefore, we investigate our method on three subsets that contain different categories.
The statistics of our dataset and corresponding attributed graphs are presented in Table~\ref{table:dataset}. Recall that the text graph consists of paper vertices and word vertices, and thus contains one additional class because all word vertices are labeled as a new class, which is not touched during the training or testing phase.

To evaluate our model, we split the train, validation, and test data as 60\%, 20\%, and 20\%. Due to the potential for random splits to result in an easier task for our model, we ran the experiments five times using random seed IDs from 0 to 4 and reported the mean values with corresponding standard deviations, mean (std). We evaluate the classification performance by accuracy and weighted f1 score. Accuracy is a common metric on classification tasks, whereas the weighted f1 score provides a balanced measure of the class-imbalanced dataset.

\begin{table*}[h]
\small
\centering
\setlength{\tabcolsep}{2.5pt}
\scalebox{0.9}{
\begin{tabular}{ccccccc}
\toprule
%\multicolumn{1}{c}{\multirow{2}{*}{\bf Models}} 
& \multicolumn{2}{c}{\bf $Data_{Nov23}$} & \multicolumn{2}{c}{\bf $Data_{Jan24}$} & \multicolumn{2}{c}{\bf $Data_{subset}$} \\
\cmidrule(lr){2-3} \cmidrule(lr){4-5} \cmidrule(lr){6-7}
& Accuracy & Weighted-F1 & Accuracy & Weighted-F1 & Accuracy & Weighted-F1 \\
\midrule
{\bf BERT~\cite{kenton2019bert}} & 30.43 (18.45) & 25.70 (19.91) & 43.45 (18.84) & 41.50 (22.31) & 58.74 (6.87) & 57.51 (6.94) \\
{\bf RoBERTa~\cite{liu2019roberta}} & 41.74 (20.32) & 39.17 (22.84) & 35.86 (17.53) & 27.23 (23.67) & 25.93 (12.17) & 17.02 (15.16) \\
{\bf DistilBERT~\cite{sanh2019distilbert}} & {\bf 57.39 (8.87)} & {\bf 55.59 (10.66)} & {\bf 53.10 (2.76)} & {\bf 52.07 (4.47)} & {\bf 59.26 (7.41)} & {\bf 58.15 (8.82)} \\
{\bf XLNet~\cite{yang2019xlnet}} & 25.22 (16.82) & 21.59 (20.54) & 27.59 (14.30) & 21.59 (15.39) & 28.52 (9.37) & 21.51 (9.65) \\
{\bf Electra~\cite{clark2019electra}} & 23.04 (4.76) & 19.06 (4.06) & 44.83 (7.23) & 42.01 (8.39) & 20.01 (6.87) & 12.03 (8.45) \\
{\bf Albert~\cite{lan2019albert}} & 11.30 (8.06) & 4.85 (7.21) & 15.17 (4.68) & 5.14 (2.87) & 20.74 (9.83) & 11.41 (11.15) \\
{\bf BART~\cite{lewis2020bart}} & 51.30 (17.48) & 50.30 (17.62) & 51.72 (3.08) & 50.62 (2.79) & 58.25 (8.11) & 57.68 (8.90) \\
{\bf DeBERTa~\cite{he2020deberta}} & 24.78 (8.06) & 19.61 (10.36) & 26.21 (11.03) & 20.92 (14.25) & 25.93 (10.73) & 24.30 (10.52) \\
{\bf Llama2~\cite{touvron2023llama}} & 14.48 (8.72) & 4.77 (4.35) & 19.22 (5.90) & 6.03 (4.21) & 23.45 (8.72) & 12.59 (7.23) \\
\bottomrule
\end{tabular}
}
\caption{Evaluation of fine-tuning pre-trained language models on the text data across three subsets.}
\label{tab:lm}
\end{table*}

\subsection{Leveraging Graph Structure Information for Taxonomy Classification}

We investigate whether leveraging the graph structure information can help better classify the papers to their corresponding categories in the proposed taxonomy.
In this experiment, we construct the attributed graphs based on the text data (including the title and summary) and the relationship of the co-authorship and co-category. To examine the generalization of \grl, we employ GCN~\cite{kipf2016semi} as a backbone \gnn\ on various graph structures across three subsets.
According to Table~\ref{tab:gnn}, \gnns\ fail to learn graph representation on both the text graph and the co-author graph.
For the text graph, we argue that the degradation may be caused by excessively similar words in the summary of survey papers. When constructing the text graph, these word vertices connect with many paper vertices, resulting in the paper vertices being less distinguishable.
For the co-author graph, we conjecture that it is challenging to categorize papers solely based on the sparse co-authorship in this dataset. Furthermore, we observe that some co-authorships come from a common mentor in the same lab whereas two first authors work on the survey papers in two distinct categories. These reasons weaken the effectiveness of using graph structure information. \gnns, in contrast, are very reliable (evaluated by both accuracy and weighted F1 score) in most co-category graphs.

\paragraph{Ablation Analysis}
We further examine the graph structures of co-category graphs by conducting ablation studies.
First, according to Figure~\ref{fig:cat_dist}, most papers are assigned as \texttt{cs.CL} and \texttt{cs.AI} in the arXiv categories. Thus, we study how the categories, \texttt{cs.CL} and \texttt{cs.AI}, affect the performance by muting these two categories in a combinatorial manner. In Table~\ref{tab:gnn}, we observe that \gnns\ can maintain a comparable performance after removing either \texttt{cs.CL} or \texttt{cs.AI}. However, the performance dramatically drops after removing both categories. This is possible since most node connections are significantly sparsified after these two categories are removed. Even though both \texttt{cs.CL} and \texttt{cs.AI} do not directly map to the existing classes, either one can connect the nodes and further strengthen the message-passing in \gnns, allowing \gnns\ to learn better node representations.

We visualize co-category graphs in $Data_{Jan24}$ in Figure~\ref{fig:cocat_graphs}. The visualization indicates that most nodes are clustered well even if we remove the category either \texttt{cs.CL} or \texttt{cs.AI}. However, after removing these two categories simultaneously, we observe that node classifications gradually become disordered and several nodes are then isolated. This visualization illustrates the effectiveness of \grl.

We further visualize GCNs' hidden representation on the above co-category graphs in $Data_{Jan24}$ in Figure~\ref{fig:cocat_embs}. The figures show that the nodes are well-classified in the hidden space even if either the category \texttt{cs.CL} or \texttt{cs.AI} is removed. However, the distribution of nodes tends to become chaotic when both of these two categories are removed simultaneously, shown in Table~\ref{tab:gnn}.

For completeness, we conducted another ablation study to examine how the categories, \texttt{cs.IR}, \texttt{cs.RO}, and \texttt{cs.SE}, affect the classification performance as their names are similar to that of some classes in our proposed taxonomy (recall that our proposed taxonomy is not based on the arXiv categories). According to Table~\ref{tab:gnn}, the classification performances are well-maintained no matter which category is removed, whereas removing \texttt{cs.SE} does slightly change the results (highlighted by gray color). We argue that the results are reasonable since these removals only drop a small number of edges and don't break the topological relationships in the graph. Overall, these studies verify that leveraging the graph structure information in co-category graphs can positively contribute to the taxonomy classification.

\subsection{Fine-tuning Pre-trained Language Models}
After verifying the effectiveness of \grl, we continue to investigate whether \grl\ can transcend fine-tuning pre-trained language models on the text data across three subsets in the taxonomy classification task. To preprocess the text data, we follow the same setup as the cleaning process for text graphs. In this fine-tuning paradigm, we use various transformer-based~\cite{vaswani2017attention} pre-trained language models as competing models, such as BERT~\cite{kenton2019bert}, which learns bidirectional representations and significantly enhances performance across a wide range of contextual understanding tasks. The results in Table~\ref{tab:gnn} and Table~\ref{tab:lm} gave us an affirmative answer to the superiority of \grl. In Table~\ref{tab:lm}, we further observe that medium-size language models, such as DistilBERT~\cite{sanh2019distilbert}, work better on smaller text data. However, the performance may dramatically drop when the model size is too small, such as Albert~\cite{lan2019albert} or too large, such as Llama2~\cite{touvron2023llama}. We conjecture that a smaller pre-trained model may be more sensitive to the domain shift issue (our dataset has distinct class distributions compared to that of the dataset used to pre-train the language models), whereas a large pre-trained model may suffer overfitting issues when it is fine-tuned on a smaller text data.

\paragraph{Fine-tuning with Weak Labels}

\begin{figure}[h]
  \centering
  \includegraphics[width=0.9\linewidth]{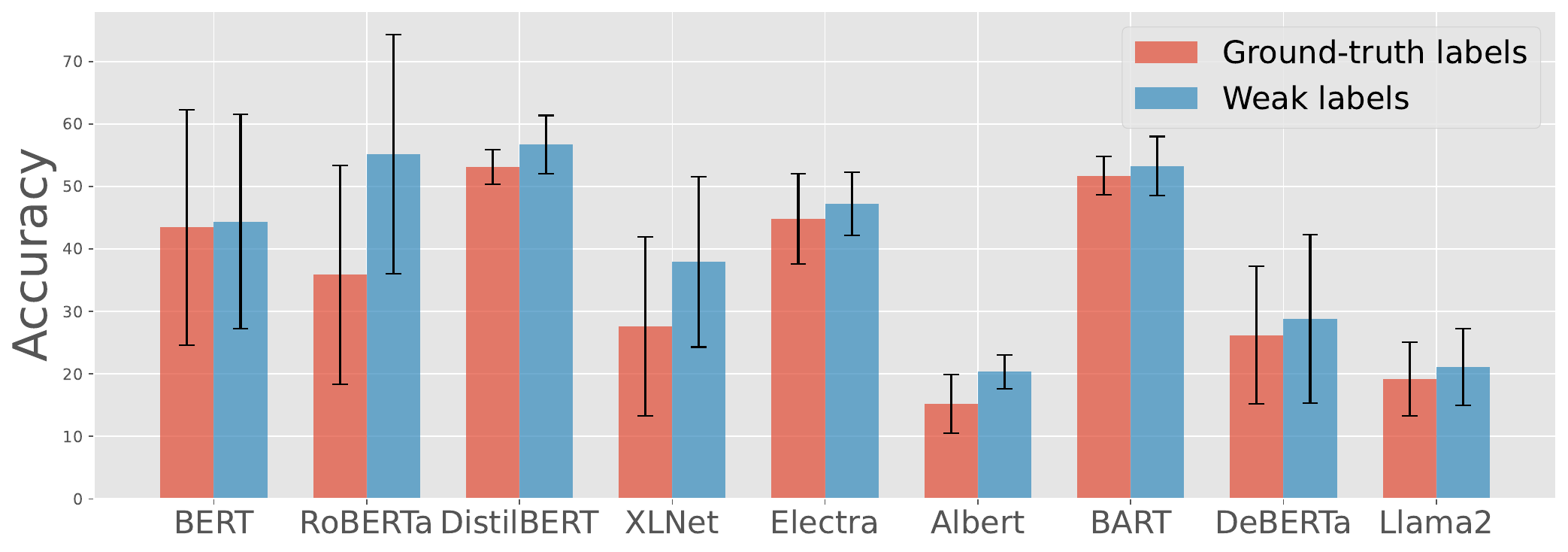}
  \caption{Comparison of fine-tuning pre-trained language models using ground-truth labels and weak labels. Whiskers denote standard deviation.}
\label{fig:lm_weak}
\end{figure}

The above experiments confirmed that smaller ("weaker") \gnns\ can surpass larger ("stronger") language models in the taxonomy classification task. Recently, \citet{burns2023weak}. verified that training stronger models with pseudo labels, a.k.a. weak labels, generated by weaker models can enable the stronger models to achieve comparable performance as closely as those trained with ground-truth labels. In this experiment, we first generate weak labels by GCN on co-category graphs and then fine-tune pre-trained language models with weak labels. We present the results on $Data_{Jan24}$ in Figure~\ref{fig:lm_weak} as an example. The results indicate that performance achieved through training with weak labels can surpass that of training with ground-truth labels. One possible reason is that training the model using noisy labels with a low noise ratio can be equivalent to a kind of regularization, improving the classification results~\cite{zhuang2022robust}.\footnote{Weak labels can be regarded as a kind of noisy label.}

This experiment demonstrates that leveraging weak labels generated by smaller models may effectively enhance the performance of larger models. This is one of the applications related to "weak-to-strong generalization"~\cite{burns2023weak}.

\subsection{LLM Zero-shot/Few-shot Classification and Human Evaluation}

\begin{table}[h]
\small
\centering
\begin{tabular}{ccc}
\toprule
 & {\bf Accuracy} & {\bf Weighted-F1} \\
\midrule
{\bf Human} & \textcolor{red}{\bf 58.73 (19.16)} & \textcolor{red}{\bf 59.50 (19.13)} \\
{\bf Claude w.o. hints} & 11.61 (1.27) & 12.66 (0.14) \\
{\bf Claude w. hints} & 10.27 (3.15) & 12.81 (2.10) \\
{\bf GPT 3.5 w.o. hints} & 47.32 (3.25) & 43.21 (4.33) \\
{\bf GPT 3.5 w. hints} & {\bf 53.57 (2.81)} & {\bf 53.16 (3.13)} \\
{\bf GPT 4 w.o. hints} & 29.76 (7.22) & 26.91 (9.66) \\
{\bf GPT 4 w. hints} & 33.04 (5.57) & 27.78 (7.76) \\
\bottomrule
\end{tabular}
\caption{Evaluation of zero-shot and few-shot classification capabilities of large language models, Claude, GPT 3.5, and GPT 4. We compare these results with those from human recognition.}
\label{tab:llm}
\end{table}

\begin{table}[h]
\small
\centering
\begin{tabular}{cccccc}
\toprule
{\bf Major} & CS & DS & Security & Math & Chemistry  \\
\midrule
{\bf \#Students} & 9 & 4 & 2 & 2 & 1 \\
\bottomrule
\end{tabular}
\caption{The number of students in each major. CS, DS, and Security denote the major in Computer Science, Data Science, and Cyber-security, respectively.}
\label{tab:majors}
\end{table}

In this experiment, we evaluate zero-shot/few-shot classification capabilities of \llms\, Claude~\cite{bai2022constitutional}, GPT 3.5~\cite{brown2020language}, and GPT 4~\cite{achiam2023gpt}, on the text data, which contains both title and summary, on $Data_{Nov23}$ as an example.
We also compare the results with human participants. We recruited 18 students across five different majors in a graduate-level course. The number of students in each major is shown in Table~\ref{tab:majors}. Each participant was given the titles and abstracts of survey papers and was asked to assign a category to each paper from our taxonomy. We present the mean value with the corresponding standard deviation in Table~\ref{tab:llm}. For the \llms, we ran the experiments five times.
The standard deviation in human recognition is relatively large as some students do not have a strong technical background so they perform worse in this test.
Among the \llms, GPT 3.5 outperforms the other two models given that all models have not seen the data before (zero-shot). We further provide some hints to the models before classification (few-shot). For example, we release the keywords of the class "Trustworthy" to the models before classification. In this setting, both GPT 3.5 and GPT 4 can achieve higher accuracy and a weighted F1 score after obtaining some hints.
In brief, \grl\ can outperform all three \llms\ and human recognition, whereas these \llms\ couldn't surmount human recognition, which reveals that \llms\ still have much room to improve in taxonomy classification.

\section{Conclusion}
In this work, we aim to develop a method to automatically assign survey papers about Large Language Models (\llms) to a taxonomy. To achieve this goal, we first collected the metadata of 144 \llm\ survey papers and proposed a new taxonomy for these papers. We further explored three paradigms to classify survey papers into the categories in the proposed taxonomy. After investigating three types of attributed graphs, we observed that leveraging graph structure information on co-category graphs can significantly help the taxonomy classification.
Furthermore, our analysis validates that graph representation learning outperforms pre-trained language models’ fine-tuning, zero-shot/few-shot classifications using \llms, and even surpasses an average human recognition level.
Last but not least, our experiments indicate that fine-tuning pre-trained language models using weak labels, which are generated by a weaker model, such as GCN, can be more effective than using ground-truth labels, revealing the potential for weak-to-strong generalization in the taxonomy classification task.

\paragraph{Limitations \& Future Work}
Constructing a graph structure may encounter certain constraints. For instance, we build co-category graphs based on the arXiv categories. When papers come from distinct fields, such as biology, physics, and computer science, the graph structure may be very sparse, weakening the effectiveness of \grl.

In the future, our primary motivation extended from this study is to tailor GPT-based applications to assist readers in understanding survey papers more effectively. We also plan on further exploring the weak-to-strong generalization which could potentially have many important applications. 

% Entries for the entire Anthology, followed by custom entries
\bibliography{anthology,custom,paperpile}
\bibliographystyle{acl_natbib}

\begin{figure*}[t]
  \centering
  \raisebox{0.4\height}{\rotatebox{90}{$Data_{Nov23}$}}
  %\hfill
  \begin{subfigure}{0.235\linewidth}
    \centering  % include the 1st image
    \includegraphics[width=\textwidth]{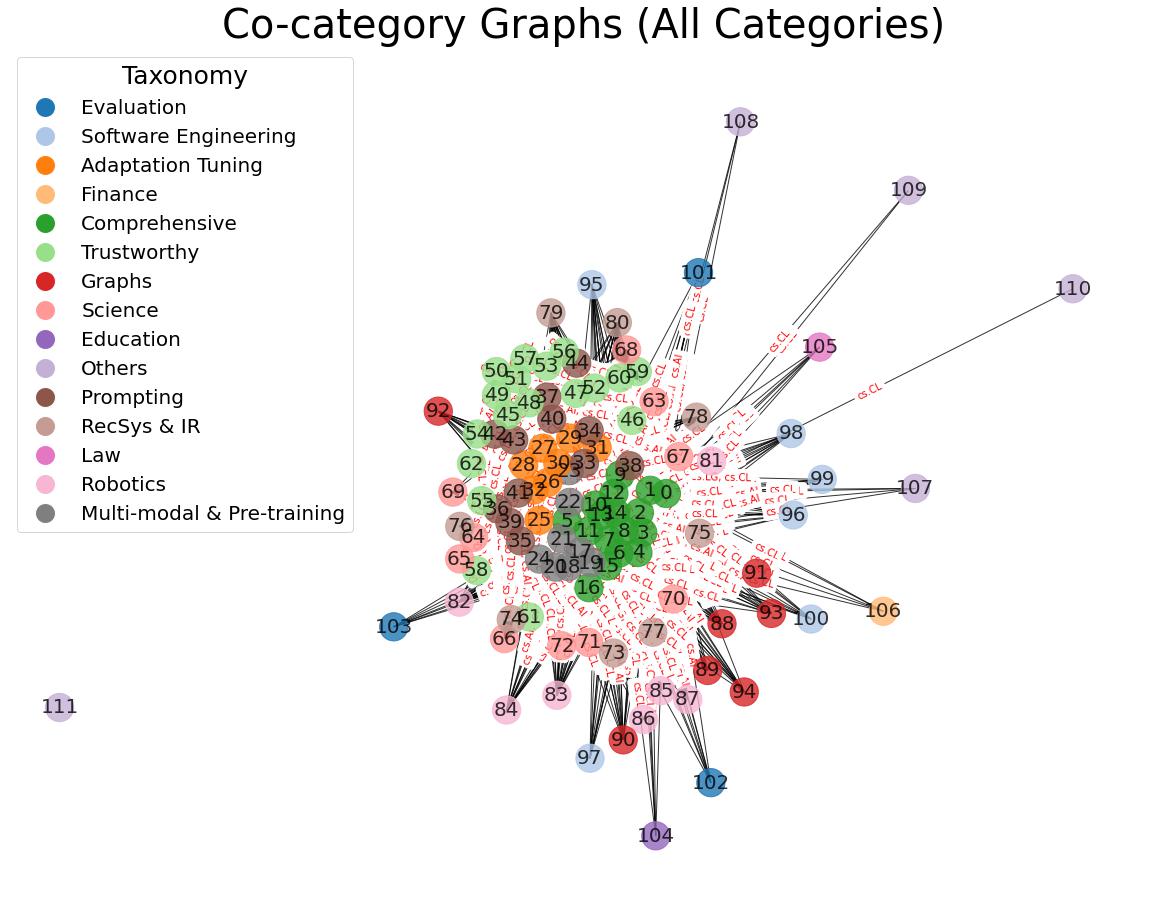}
  \end{subfigure}
  %\hfill
  \begin{subfigure}{0.235\linewidth}
    \centering  % include the 3rd image
    \includegraphics[width=\textwidth]{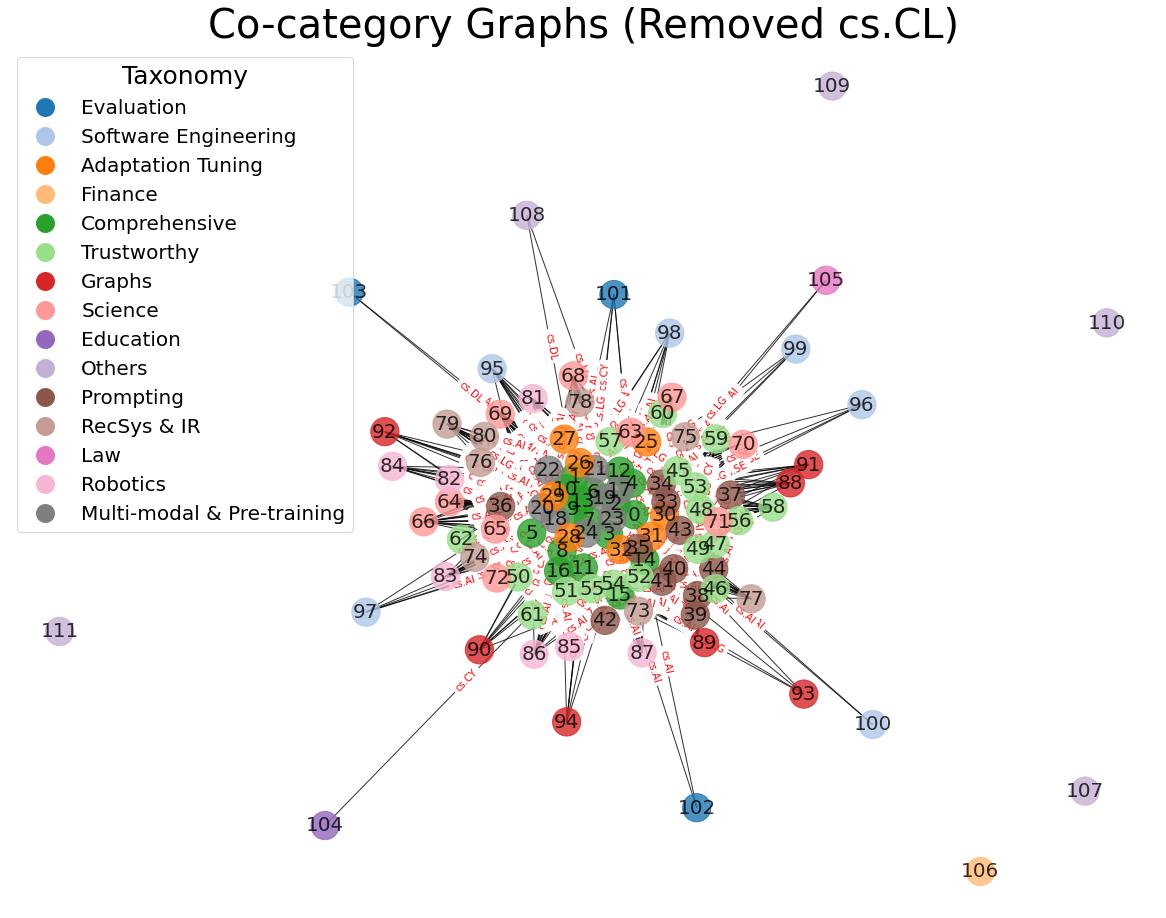}
  \end{subfigure}
  %\hfill
  \begin{subfigure}{0.235\linewidth}
    \centering  % include the 4th image
    \includegraphics[width=\textwidth]{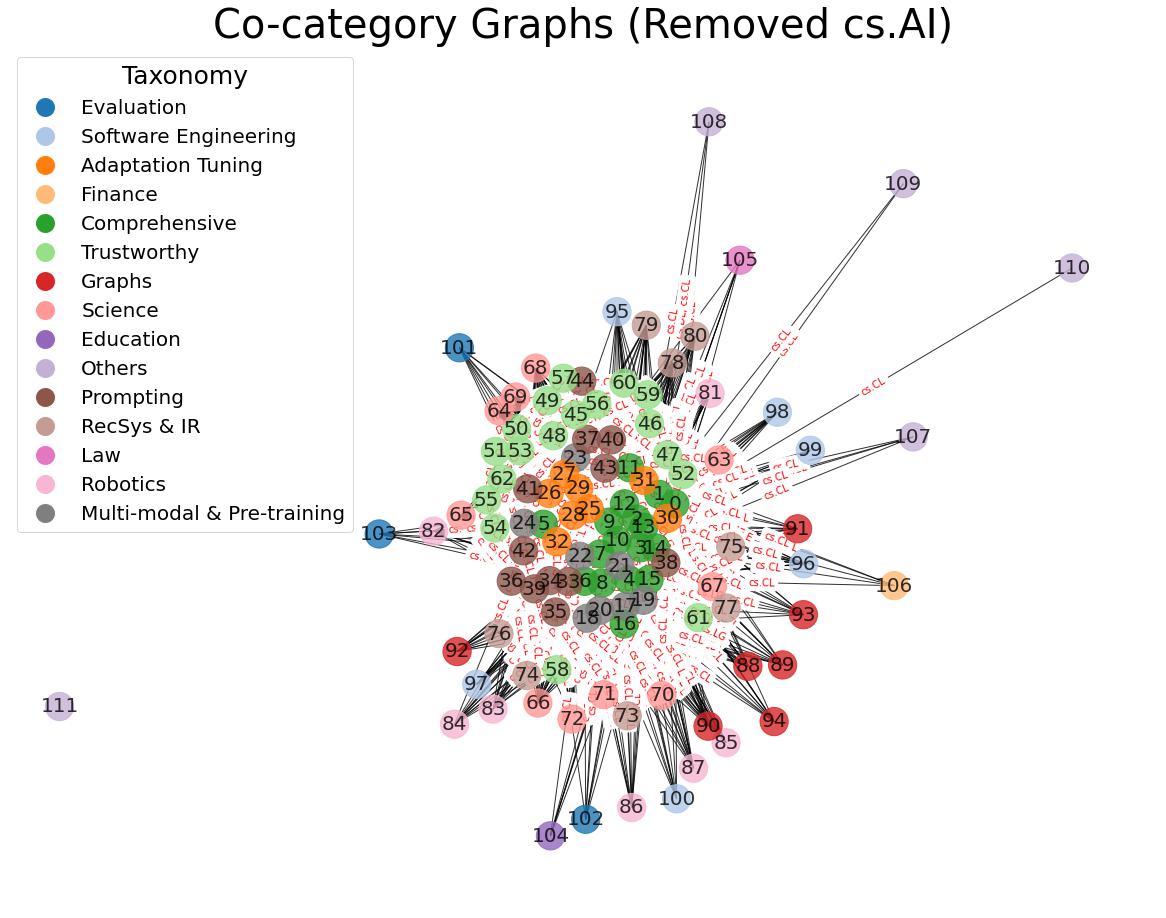}
  \end{subfigure}
  %\hfill
  \begin{subfigure}{0.235\linewidth}
    \centering  % include the 2nd image
    \includegraphics[width=\textwidth]{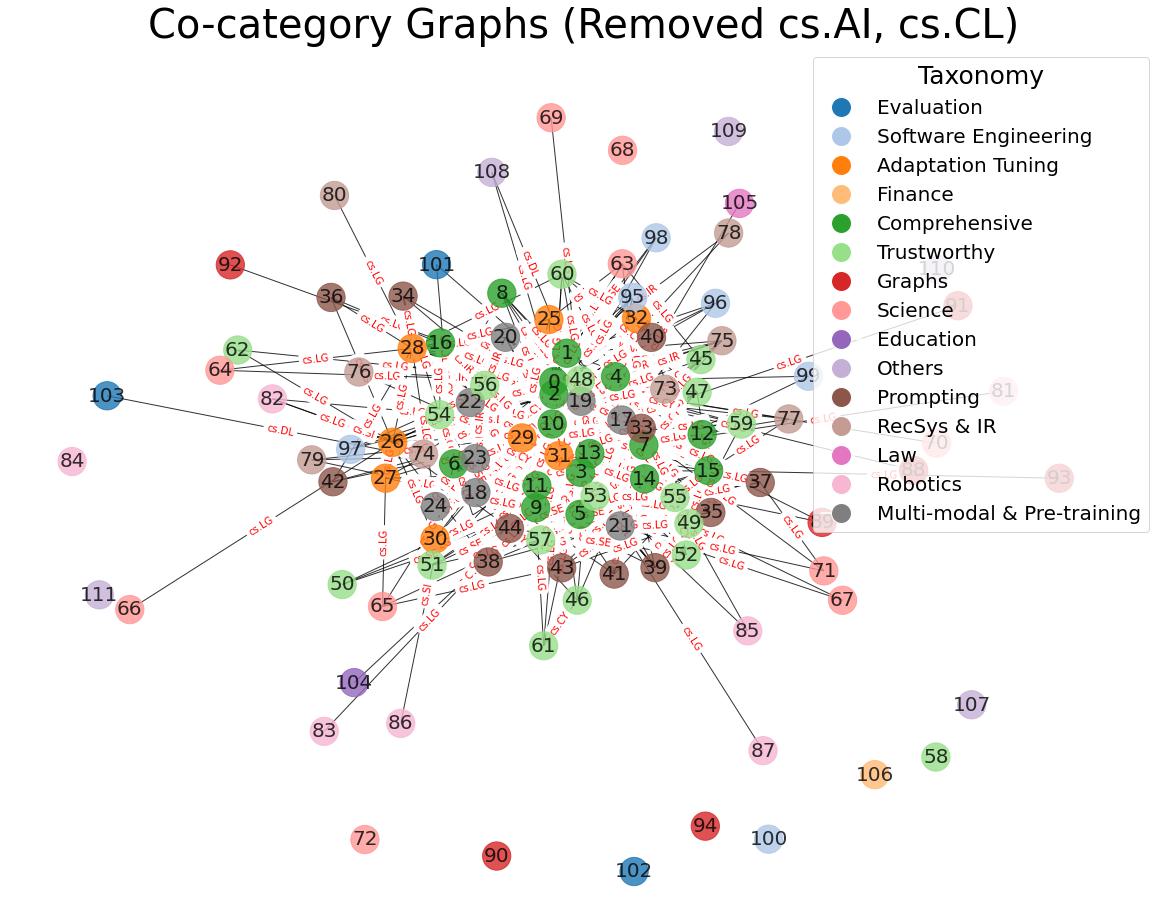}
  \end{subfigure}
  % -----
  \centering
  \raisebox{0.4\height}{\rotatebox{90}{$Data_{Subset}$}}
    %\hfill
  \begin{subfigure}{0.235\linewidth}
    \centering  % include the 1st image
    \includegraphics[width=\textwidth]{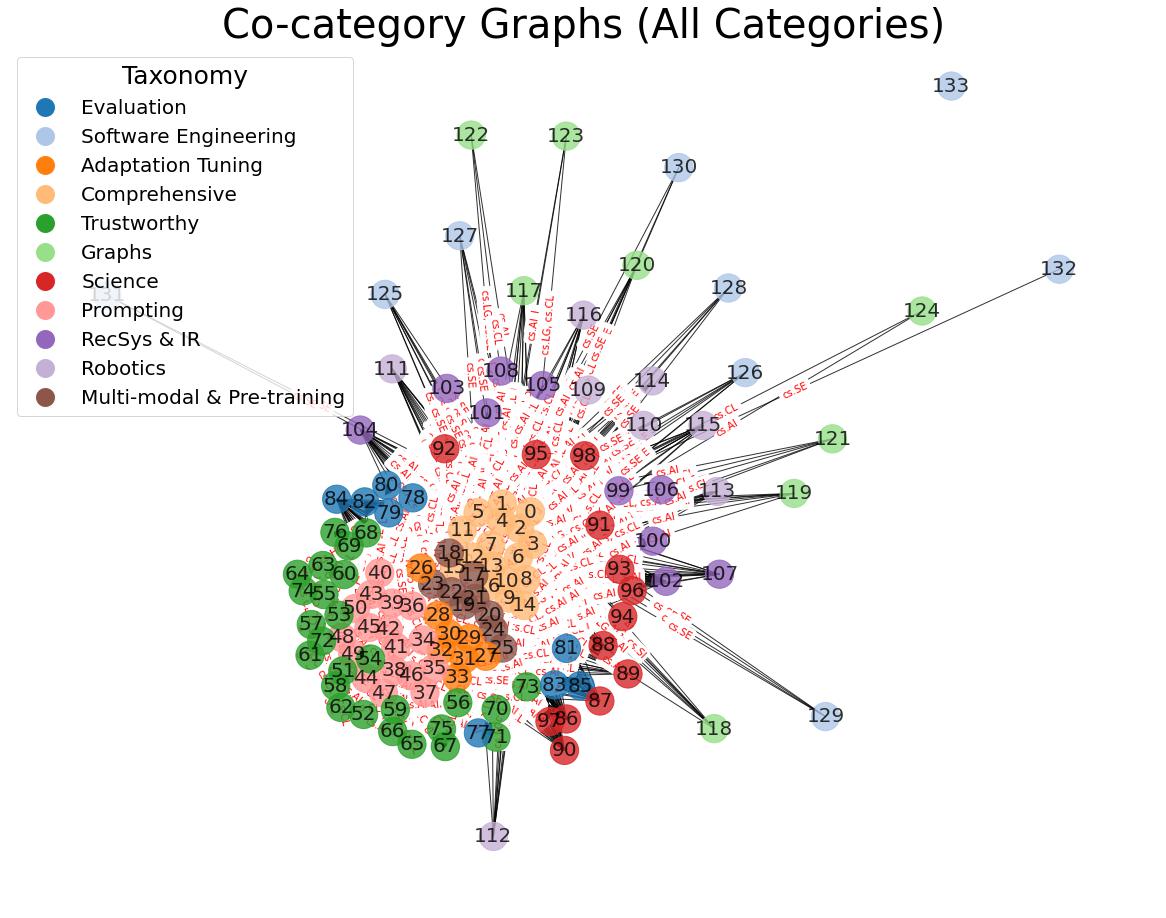}
  \end{subfigure}
  %\hfill
  \begin{subfigure}{0.235\linewidth}
    \centering  % include the 3rd image
    \includegraphics[width=\textwidth]{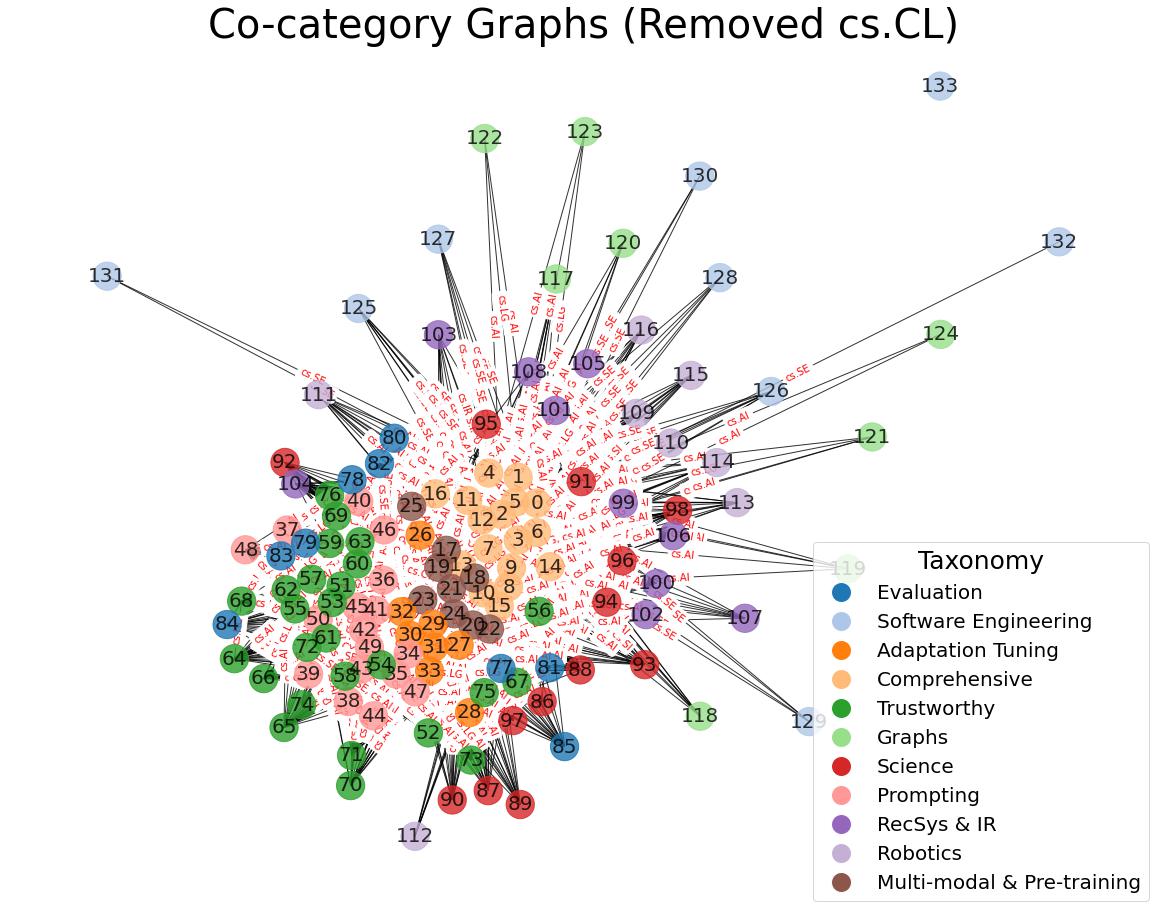}
  \end{subfigure}
  %\hfill
  \begin{subfigure}{0.235\linewidth}
    \centering  % include the 4th image
    \includegraphics[width=\textwidth]{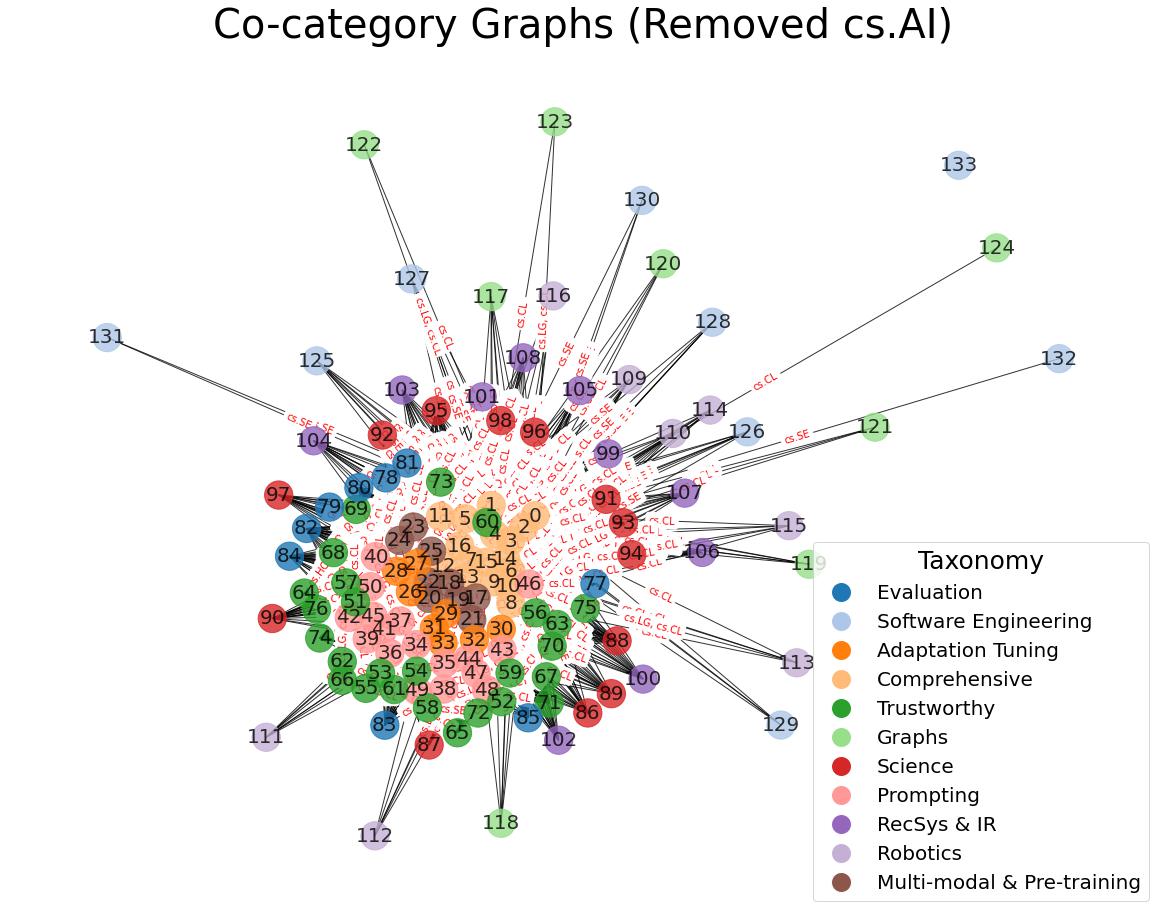}
  \end{subfigure}
  %\hfill
  \begin{subfigure}{0.235\linewidth}
    \centering  % include the 2nd image
    \includegraphics[width=\textwidth]{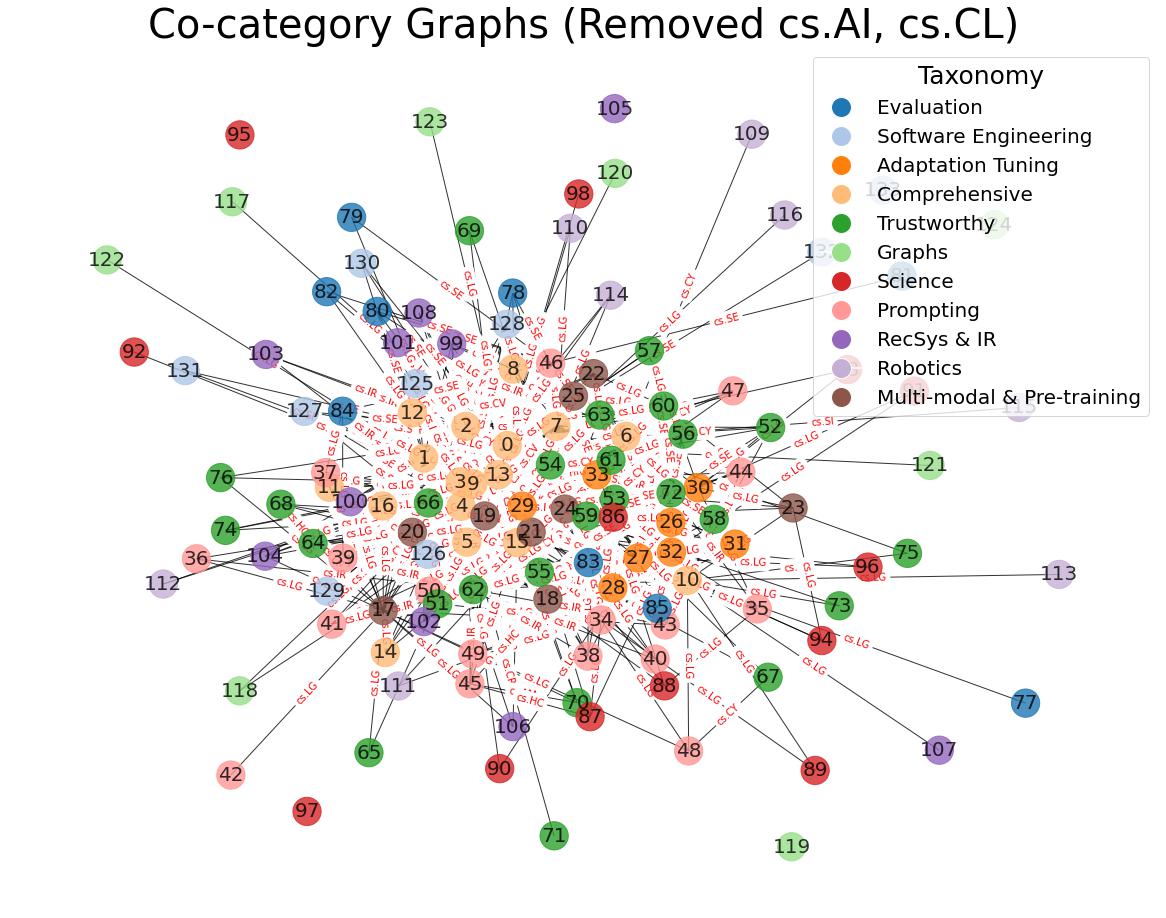}
  \end{subfigure}
\caption{Additional visualization of co-category graphs (extended from Figure~\ref{fig:cocat_graphs}).}
\label{fig:cocat_graphs_apdx}
\end{figure*}

\begin{figure}[t]
  \centering
  \begin{subfigure}{0.48\textwidth}
    \centering
    \raisebox{0.4\height}{\rotatebox{90}{\scriptsize $Data_{Nov23}$}}
    \includegraphics[width=0.9\textwidth]{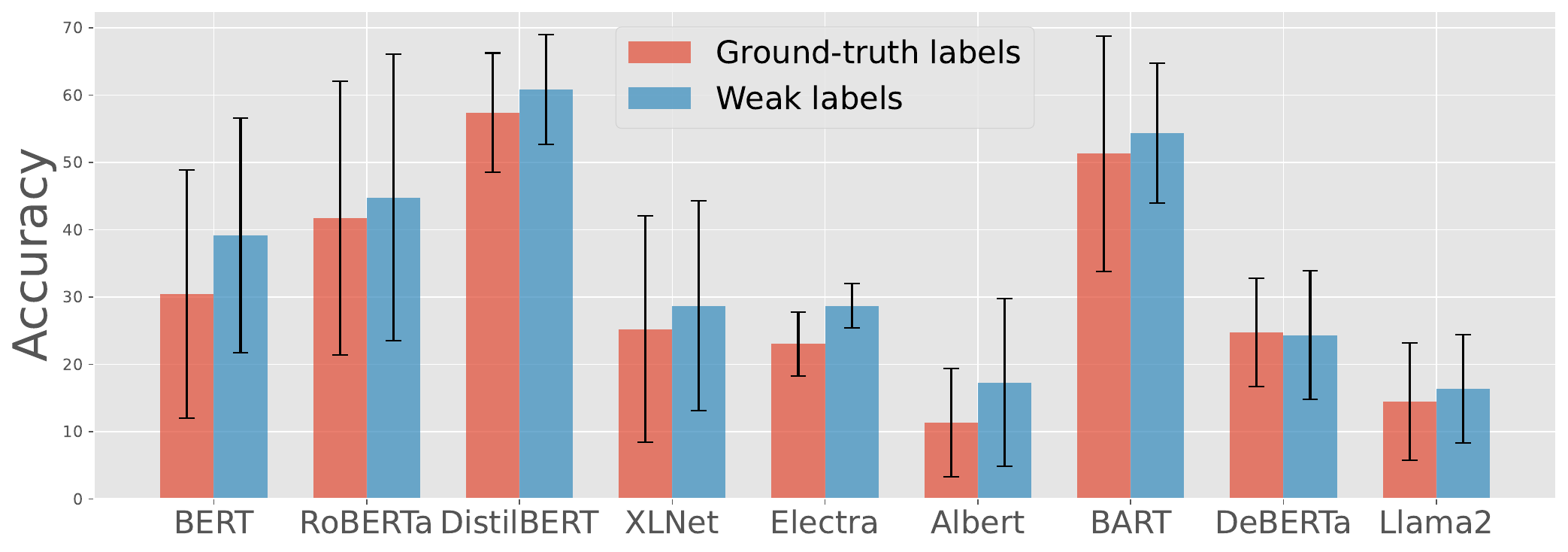}
  \end{subfigure}
  \begin{subfigure}{0.48\textwidth}
    \centering
    \raisebox{0.4\height}{\rotatebox{90}{\scriptsize $Data_{subset}$}}
    \includegraphics[width=0.9\textwidth]{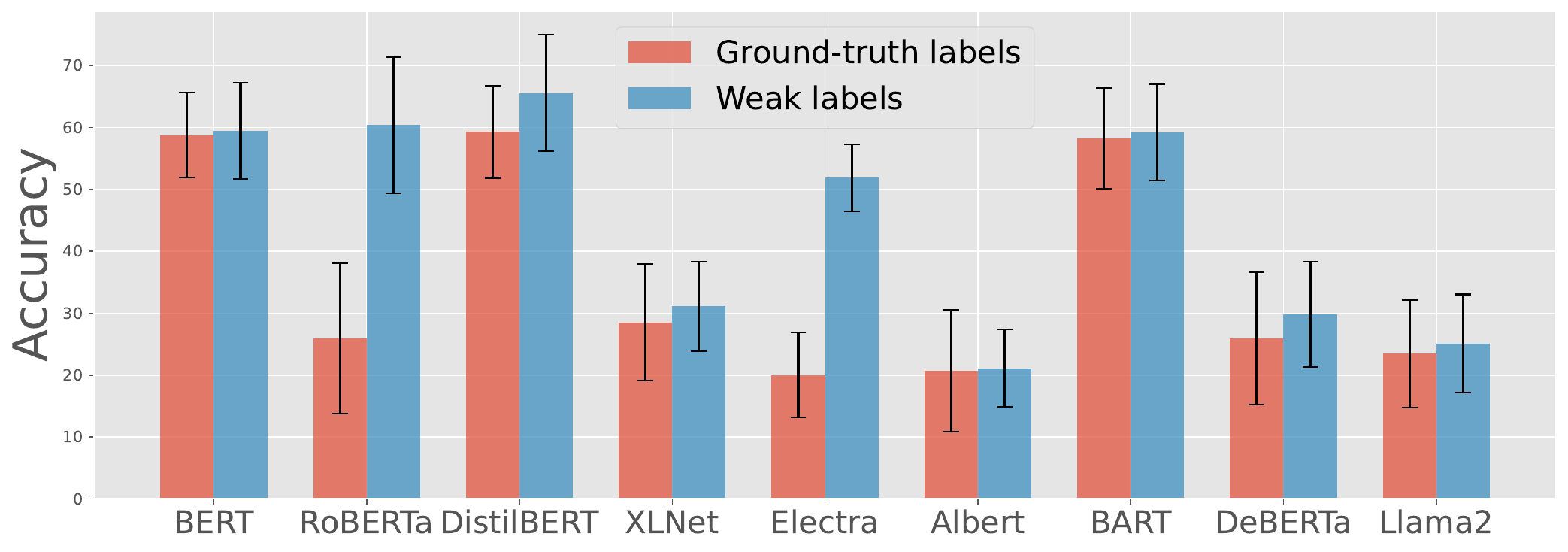}
  \end{subfigure}
\caption{Additional comparison in the fine-tuning paradigm using weak labels (extended from Figure~\ref{fig:lm_weak}).}
\label{fig:lm_weak_apdx}
\end{figure}

\appendix
\section{APPENDIX}
\label{sec:appendix}

In the appendix, we present the \gnn\ and pre-trained language models' hyper-parameters and the hardware and software. We also include the additional comparison results about fine-tuning using weak labels and additional visualization of co-category graphs.

\paragraph{Hyper-parameters and Settings}
We employ a two-layer GCN~\cite{kipf2016semi} with 200 hidden units and a ReLU activation function as the backbone \gnn\ to examine the effectiveness of \grl. The \gnn\ is trained by the Adam optimizer with a learning rate, $1 \times 10^{-2}$ for both co-author graphs and co-category graphs and $2 \times 10^{-2}$ for text graphs, and converged within 500 training epochs on all subsets. The dropout rate is 0.5.

\begin{table}[h]
%\scriptsize
\small
\centering
\begin{tabular}{cc}
\toprule
\multicolumn{1}{c}{\bf Language Models} & {\bf Model Size} \\
\midrule
{\bf BERT~\cite{kenton2019bert}} & 109.49M \\
{\bf RoBERTa~\cite{liu2019roberta}} & 124.66M \\
{\bf DistilBERT~\cite{sanh2019distilbert}} & 66.97M \\
{\bf XLNet~\cite{yang2019xlnet}} & 117.32M \\
{\bf Electra~\cite{clark2019electra}} & 109.49M \\
{\bf Albert~\cite{lan2019albert}} & 11.70M \\
{\bf BART~\cite{lewis2020bart}} & 140.02M \\
{\bf DeBERTa~\cite{he2020deberta}} & 139.20M \\
{\bf Llama2~\cite{touvron2023llama}} & 6.61B \\
\bottomrule
\end{tabular}
\caption{Model size of the pre-trained language models. For example, BERT has 109.49 million parameters.}
\label{tab:lm_size}
\end{table}

We fine-tune the pre-trained language models using the Adam optimizer with a $1 \times 10^{-4}$ learning rate. We chose the batch size of 8 for the Llama2 and fixed the batch size of 16 for the rest of the models. We implement the pre-trained language models using HuggingFace packages (we choose the base version for all models) and report the model size in Table~\ref{tab:lm_size}. All models are tuned with 30 epochs.

\paragraph{Hardware and Software}
The experiment is conducted on a server with the following settings:
\begin{itemize}[itemsep=-1mm]
  \item Operating System: Ubuntu 22.04.3 LTS
  \item CPU: Intel Xeon w5-3433 @ 4.20 GHz
  \item GPU: NVIDIA RTX A6000 48GB
  \item Software: Python 3.11, PyTorch 2.1, HuggingFace 4.31, dgl 1.1.2+cu118.
\end{itemize}

\paragraph{Computational Budgets}
Based on the above computing infrastructure and settings, computational budgets in our experiments are described as follows.
The experiment presented in Table~\ref{tab:gnn} can be reproduced within one hour. The experiment displayed in Table~\ref{tab:lm} may take 93 hours to complete. Due to limited GPU memory, we implemented Llama2 using the CPU. This consumes around 90 hours in total. The experiment shown in Table~\ref{tab:llm} (excluding human recognition) can be finished in one hour.

\paragraph{Additional Visualization of Co-category Graphs}
Besides visualizing four graph structures in $Data_{Jan24}$ in Figure~\ref{fig:cocat_graphs}, we additionally present the visualization of four corresponding co-category graphs in both $Data_{Nov23}$ and $Data_{subset}$ in Figure~\ref{fig:cocat_graphs_apdx}. The visualization verifies the generalization of \grl\ across three subsets.

\paragraph{Additional Comparison of Fine-tuning Using Weak Labels}
Besides the results in Figure~\ref{fig:lm_weak}, we supplement the comparisons on both $Data_{Nov23}$ and $Data_{subset}$ in Figure~\ref{fig:lm_weak_apdx}. The comparisons across nice pre-trained language models further validate the effectiveness of fine-tuning using weak labels.

\paragraph{Ethical and Broader Impacts}
We confirm that we fulfill the author's responsibilities and address the potential ethical issues. In this work, we aim to help researchers quickly and better understand a new research field. Many researchers in academia or industry may potentially benefit from our work.

\paragraph{Statement of Data Privacy}
Our dataset contains the authors' names in each paper. This information is publicly available so the collection process doesn't infringe on personal privacy.

\paragraph{Disclaimer Regarding Human Subjects Results}
In Table~\ref{tab:llm}, we include partial results with human subjects. We already obtained approval from the Institutional Review Board (IRB). The protocol number is IRB24-056. We recruited volunteers from a graduate-level course. Before the assessment, we have disclaimed the potential risk (our assessment has no potential risk) and got consent from participants.

\end{document}